\documentclass{article} 
\usepackage{iclr2026_conference,times}

\usepackage{amsmath,amsfonts,bm}
\usepackage[acronyms,shortcuts]{glossaries}

\def\eqref#1{equation~\ref{#1}}

\def\1{\bm{1}}

\DeclareMathAlphabet{\mathsfit}{\encodingdefault}{\sfdefault}{m}{sl}
\SetMathAlphabet{\mathsfit}{bold}{\encodingdefault}{\sfdefault}{bx}{n}

\newcommand{\methodname}{\texttt{DiffVax}}

\newcommand{\image}{\mathbf{I}}
\newcommand{\mask}{\mathbf{M}}

\newcommand{\diffusion}[1]{\text{SD}(#1)}
\newcommand{\unet}[1]{f(#1;\theta)}
\newcommand{\unetparameters}{\theta}
\newcommand{\prompt}{\mathcal{P}}
\newcommand{\editedimage}{\image_{\mathrm{im,edit}}}
\newcommand{\immunizationnoise}{\epsilon_{\mathrm{im}}}
\newcommand{\immunizedimage}{\image_{\mathrm{im}}}
\newcommand{\dataset}{\mathcal{D}}
\newcommand{\noiseloss}{\mathcal{L}_\mathrm{noise}}
\newcommand{\editloss}{\mathcal{L}_\mathrm{edit}}
\newcommand{\totalloss}{\mathcal{L}}
\newcommand{\latent}{z}
\newcommand{\encoder}[1]{\mathcal{E}(#1)}

\newcommand{\denoiser}{\epsilon_{\theta}}

\definecolor{goodcolor}{rgb}{0.1, 0.7, 0.0}
\definecolor{mehcolor}{rgb}{0.8, 0.6, 0.0}
\definecolor{badcolor}{rgb}{0.8, 0.0, 0.0}

\definecolor{lightgrey}{rgb}{0.9, 0.9, 0.9}
\definecolor{cvprblue}{rgb}{0.21,0.49,0.74}

\glsdisablehyper
 
\newacronym{gan}{GAN}{Generative Adversarial Network}
\newacronym{uan}{UAN}{Universal Adversarial Network}
\newacronym{sam}{SAM}{Segment Anything Model}
\newacronym{uap}{UAP}{Universal Adversarial Perturbation}
\newacronym{ldm}{LDM}{Latent Diffusion Model}
\newacronym{dm}{DM}{Diffusion Model}
\newacronym{pgd}{PGD}{Projected Gradient Descent}
\newacronym{fd}{FD}{Frechet Distance}
\newacronym{cnn}{CNN}{convolutional neural network}
\newacronym{auc}{AUC}{area under the curve}
\newacronym{fc}{FC}{fully connected}

\usepackage[colorlinks=true, citecolor=blue, linkcolor=blue, urlcolor=blue]{hyperref}
\usepackage{url}
\usepackage{xcolor}         
\usepackage{graphicx}
\usepackage{amsmath}
\usepackage{algorithm}
\usepackage{algpseudocode}
\usepackage{etoc}
\usepackage{titlesec}
\usepackage{pifont}  
\usepackage{colortbl}
\usepackage{booktabs}
\usepackage[acronyms,shortcuts]{glossaries}
\usepackage{multirow}
\usepackage{caption}
\usepackage{subcaption}  

\title{DiffVax:~Optimization-Free Image Immunization Against Diffusion-Based Editing}

\author{
  \begin{tabular}{c}
    Tarik Can Ozden\footnotemark[1] \\
    \textnormal{UIUC}
  \end{tabular}
  \And
  \begin{tabular}{c}
    Ozgur Kara\thanks{Equal contribution} \\
    \textnormal{UIUC}
  \end{tabular}
  \And
  \begin{tabular}{c}
    Oguzhan Akcin \\
    \textnormal{UT Austin}
  \end{tabular}
  \And
  \begin{tabular}{c}
    Kerem Zaman \\
    \textnormal{UNC Chapel Hill}
  \end{tabular}
  \AND
  \begin{tabular}{c}
    Shashank Srivastava \\
    \textnormal{UNC Chapel Hill}
  \end{tabular}
  \And
  \begin{tabular}{c}
    Sandeep P. Chinchali \\
    \textnormal{UT Austin}
  \end{tabular}
  \And
  \begin{tabular}{c}
    James M. Rehg \\
    \textnormal{UIUC}
  \end{tabular}
}

\makeatletter
\renewcommand{\paragraph}{%
  \@startsection{paragraph}{4}%
  {\z@}{0ex \@plus 0ex \@minus .2ex}{-0.5em}%
  {\normalfont\normalsize\bfseries}%
}

\iclrfinalcopy 
\begin{document}

\maketitle

\vspace{-1em}
\begin{center}
    \captionsetup{type=figure}
    \includegraphics[width=\textwidth]{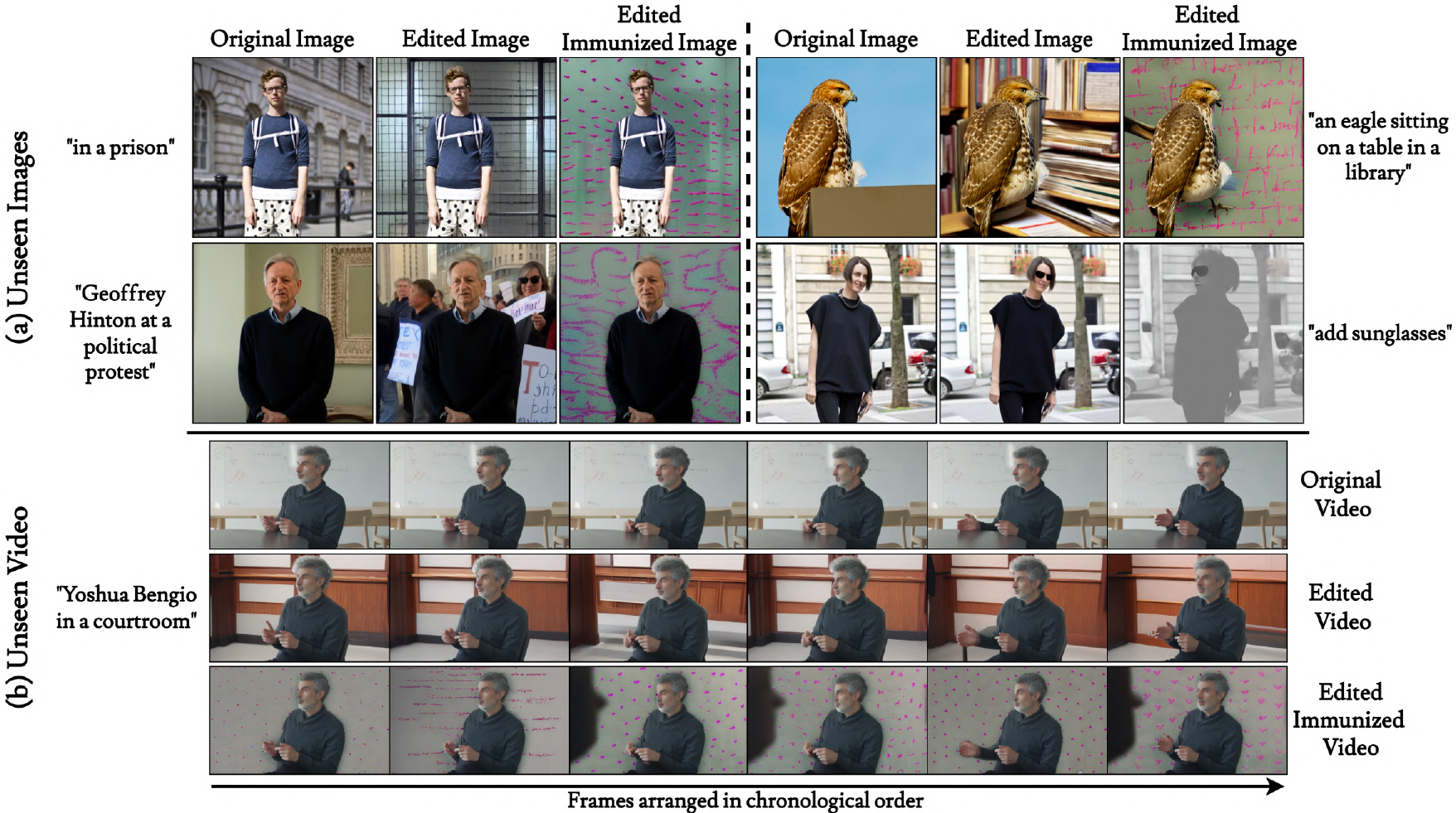} 
    \captionof{figure}{\methodname \space is an optimization-free image immunization approach designed to protect images and videos from diffusion-based editing. \methodname \space demonstrates robustness across diverse content, providing protection for both in-the-wild (a) \textit{unseen images} and (b) \textit{unseen video} content while effectively preventing edits across various editing methods, including \textit{inpainting} (illustrated with a \textit{human} in the left column and a \textit{non-human foreground object} in the right column) and \textit{instruction-based edits} (right column) with InstructPix2Pix~\citep{brooks2023instructpix2pix}. } 
    \label{fig:teaser}
\end{center}%

\begin{abstract}
Current image immunization defense techniques against diffusion-based editing embed imperceptible noise into target images to disrupt editing models. However, these methods face scalability challenges, as they require time-consuming optimization for each image separately, taking hours for small batches. To address these challenges, we introduce \methodname, a scalable, lightweight, and optimization-free framework for image immunization, specifically designed to prevent diffusion-based editing. Our approach enables effective generalization to unseen content, reducing computational costs and cutting immunization time from days to milliseconds, achieving a speedup of 250,000×. This is achieved through a loss term that ensures the failure of editing attempts and the imperceptibility of the perturbations. 
Extensive qualitative and quantitative results demonstrate that our model is scalable, optimization-free, adaptable to various diffusion-based editing tools, robust against counter-attacks, and, for the first time, effectively protects video content from editing. More details are available in our \href{https://diffvax.github.io/}{Project Webpage}.
\end{abstract}

\section{Introduction}
\label{sec:intro}

 Recent advancements in generative models, particularly diffusion models~\citep{sohl2015deep, ho2020denoising, rombach2022high}, have enabled realistic content synthesis, which can be used for various applications, such as image generation~\citep{saharia2022photorealistic,ruiz2023dreambooth,chefer2023attend,zhang2023adding,li2023gligen,mou2024t2i,bansal2023universal} and editing~\citep{brooks2023instructpix2pix,  couairon2023diffedit,hertz2023prompttoprompt,meng2022sdedit}. However, the widespread availability and accessibility of these models introduce significant risks, as malicious actors exploit them to produce deceptive, realistic content known as deepfakes~\citep{pei2024deepfakegenerationdetectionbenchmark}. Deepfakes pose severe threats across multiple domains, from political manipulation~\citep{appel2022detectionofpolitical} and blackmail~\citep{Blancaflor2024deepfake} to biometric fraud~\citep{Wojewidka2020TheDT} and compromising trust in legal processes~\citep{Delfino2022DeepfakesOT}. Furthermore, they have become tools for sexual harassment through the creation of non-consensual explicit content 
 ~\citep{KoreanDF, Deepfaked, ViceDF}. Given the widespread accessibility of diffusion models, the scale of these threats continues to grow, underscoring the urgent need for robust defense mechanisms to protect individuals, institutions, and public trust from such misuse. 





To address these challenges, one line of research has focused on deepfake detection~\citep{Naitali2023DeepfakeAG, Passos_2024} and verification methods~\citep{Hasan2019CombatingDV}, which facilitate post-hoc identification. While effective for detection, these approaches do not proactively prevent malicious editing, as they only identify it after it happens. Another branch modifies the parameters of editing models~\citep{li2024safegen} to prevent unethical content synthesis (e.g.  NSFW material); however, the widespread availability of unrestricted generative models limits its effectiveness.
A more robust defense mechanism, known as image immunization~\citep{photoguard, Lo_2024_CVPR, yeh2021attackbestdefensenullifying, ruiz2020disrupting}, safeguards images from malicious edits by embedding imperceptible adversarial perturbation.
This approach ensures that any editing attempts lead to unintended or distorted results, proactively preventing malicious modifications rather than depending on post-hoc detection. The subtlety of this protection is particularly valuable for large-scale, publicly accessible content, such as social media, where user data is especially vulnerable to malicious attacks. By uploading immunized images instead of the original ones, users can reduce the risk of misuse by malicious actors, highlighting the potential of immunization-based methods for real-world impact.


However, current immunization approaches remain inadequate, as they do not simultaneously satisfy the key requirements of an effective defense: (i) scalability for large-scale content, (ii) memory and runtime efficiency, and (iii) robustness against counter-attacks.
PhotoGuard~\citep{photoguard} (PG) embeds adversarial perturbations into target images to disrupt components of the diffusion model by solving a constrained optimization problem via projected gradient descent~\citep{madry2018towards}. Although PhotoGuard was the first immunization model targeting diffusion-based editing, it requires over 10 minutes of runtime per image and at least 15GB of memory, causing both computational and time inefficiency. To alleviate these demands, DAYN~\citep{Lo_2024_CVPR} proposes a semantic-based attack that disrupts the diffusion model's attention mechanism during editing. While this approach reduces computational load, it remains time-inefficient like PhotoGuard,  as it requires a separate optimization process for each image and cannot generalize to unseen content.  Furthermore, both approaches are vulnerable to counter-attacks, such as denoising the added perturbation or applying JPEG compression~\citep{segura2023jpeg} to the immunized image. Consequently, neither method is practical for large-scale applications, such as safeguarding the vast volume of image and video data uploaded daily on social media platforms.



\begin{figure}
    \centering
    \includegraphics[width=\linewidth]{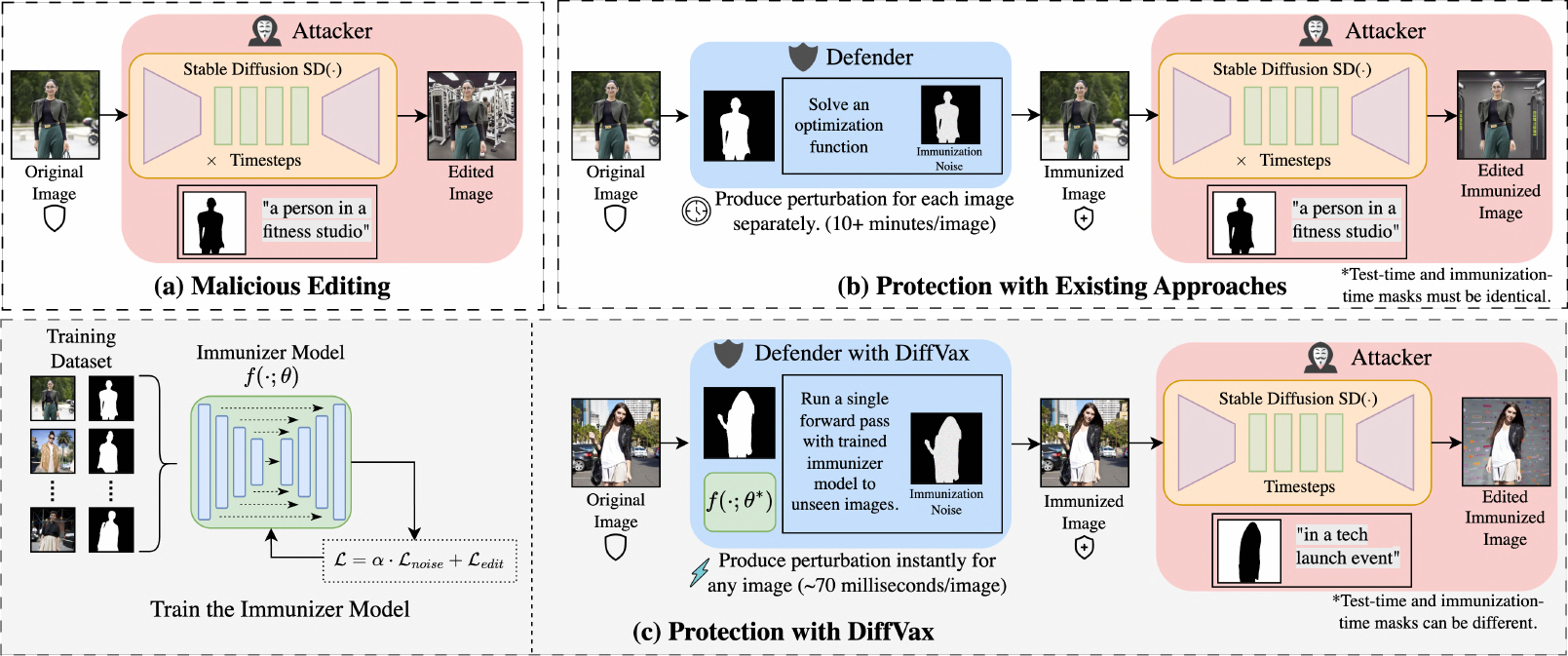}
\caption{
\textbf{\textit{Comparing \methodname\ with existing approaches.}}
\textbf{(a)} An attacker performs malicious editing on an original image.  
\textbf{(b)} Existing defenses immunize images by solving a costly optimization problem for each image individually, taking over 10 minutes per image.
\textbf{(c)} \methodname\ enables scalable protection by first training an immunizer model (green box) on a diverse dataset. Once trained, the model can immunize unseen images with a single forward pass, producing effective perturbations in approximately 70 milliseconds per image. 
}

    \label{fig:motivation}
\end{figure}

To address these challenges, we introduce \methodname, an end-to-end framework for training an ``immunizer model'' that learns how to generate imperceptible perturbations to immunize target images against diffusion-based editing (see Fig~\ref{fig:motivation}). This immunization process ensures that any attempt to edit the immunized image using a diffusion-based model fails.  \methodname\  is more effective than prior works in ensuring editing failure, and it demonstrates the feasibility and generalizability of the image-conditioned feed-forward approach to perturbation generation.

Our training process is guided by two objectives, expressed as separate terms in the loss function: (1) encouraging the model to generate an imperceptible perturbation, and (2) ensuring that any editing attempt on the immunized image fails.
Our trained immunizer operates with a single forward pass, completed within milliseconds, eliminating the need for time-intensive per-image optimization. This efficiency enables scalability to high-volume content protection. Additionally, \methodname\ enhances memory efficiency by avoiding gradient computation during inference, setting it apart from prior methods. It also exhibits robustness against common counter-attacks, such as JPEG compression and image denoising~\citep{segura2023jpeg}. In addition, our framework demonstrates superior generalization with other diffusion-based editing methods (see Fig.~\ref{fig:teaser} for examples on inpainting and instruction-based editing). Leveraging these strengths, we extend immunization to video content for the first time, achieving results previously unattainable due to the computational limitations of earlier approaches. As a result, \methodname\ satisfies all key requirements for an effective defense.


To summarize, our contributions are as follows:

\begin{itemize} 
\item 
We are the first to introduce a training framework in which the model learns to effectively immunize a given image against diffusion-based editing, drastically reducing inference time from days to milliseconds and enabling real-time protection. 

\item Thanks to its computational efficiency, our model shows promising potential as a foundational step toward immunizing video content.

\item 
Unlike prior methods that require per-image optimization and therefore cannot generalize to unseen data, our approach enables generalization to new content through a learned ``image immunizer''. 

\item \methodname \space achieves superior results with substantial degradation of the editing operation, and minimal memory requirement, demonstrating resistance to counter-attacks, making it the fastest, most cost-effective, and robust method available.



 

\end{itemize}

\section{Related Work}
\label{sec:related_works}

\paragraph{Adversarial attacks}
Adversarial attacks exploit model vulnerabilities by introducing perturbations that induce misclassification. Early gradient-based methods efficiently generated such examples via gradient manipulation~\citep{goodfellow2015explainingharnessingadversarialexamples, madry2019deeplearningmodelsresistant}, later refined to minimize perceptual distortion~\citep{carlini2017evaluatingrobustnessneuralnetworks, moosavidezfooli2016deepfoolsimpleaccuratemethod}. Generative approaches advanced these attacks by synthesizing realistic adversarial inputs~\citep{xiao2019generatingadversarialexamplesadversarial}. Subsequent work improved transferability and query efficiency using momentum and random search~\citep{dong2018boostingadversarialattacksmomentum, andriushchenko2020squareattackqueryefficientblackbox}, while ensemble-based methods strengthened robustness evaluation~\citep{croce2020reliableevaluationadversarialrobustness}. Universal perturbations~\citep{moosavidezfooli2017universaladversarialperturbations, hayes2018learninguniversaladversarialperturbations} and generative perturbation networks~\citep{poursaeed2018generative} further generalized attacks across data and models. Building on these advances, our work focuses on immunizing against diffusion-based editing, addressing its unique characteristics.

\paragraph{Preventing image editing}
The proliferation of \acp{ldm} has underscored the demand for robust immunization strategies against unauthorized image manipulation. Initial efforts focused on \ac{gan}-based models, employing adversarial perturbations to inhibit edits~\citep{yeh2021attackbestdefensenullifying, aneja2022tafimtargetedadversarialattacks}. PhotoGuard~\citep{photoguard} extended this line of work to diffusion models via encoder- and model-level perturbations but incurred substantial computational overhead due to backpropagation across multiple timesteps. To alleviate this, \citet{Lo_2024_CVPR}\footnote{Code unavailable despite request.} proposed an attention-disruption strategy that bypasses full gradient computation, though its reliance on fixed prompts limits robustness. DiffusionGuard~\citep{choi2025diffusionguard} enhances PhotoGuard by optimizing over augmented masks, yet remains computationally intensive. Similarly, PCA~\citep{guo2024greybox} proposes a grey-box attack by inducing posterior collapse in the VAE encoder to disrupt editing. Addressing instruction-guided editing, EditShield~\citep{chen2023editshield} introduces perturbations to shift latent representations, causing semantic mismatches in the edited output. Meanwhile, \citet{shih2025pixel} bypass the reliance on VAE encoders by proposing a feature-based attack effective against pixel-domain diffusion models. Other approaches, including Mist~\citep{mist}, AdvDM~\citep{advdm}, SDS~\citep{sds}, and Glaze~\citep{shawn2023glaze}, target text-to-image diffusion or fine-tuned models, but exhibit high computational demands and limited resilience to adaptive attacks. In contrast, \methodname\ introduces a model-agnostic immunizer that generalizes to unseen data via a single forward pass. Furthermore, we present, for the first time, promising results in the direction of video immunization.

\paragraph{Diffusion-based image editing}
Diffusion models have emerged as powerful tools for image editing tasks such as inpainting~\citep{wang2023stylediffusion, andreas2022repaint, zhang2023towards}, style transfer~\citep{wang2023stylediffusion, chong2024dragon, yang2023paint, hertz2023delta}, and text-guided transformations~\citep{brooks2023instructpix2pix, lin2024text, ravi2023preditor}, by conditioning on prompts or image regions. Edits are guided through attention manipulation~\citep{guarav2023pixtopix} and multi-step noise prediction. Approaches include both training-based~\citep{guillaume2023diffedit, gwanghyun2022difclip} and training-free methods~\citep{mokady2023null, daiki2023negative} requiring minimal fine-tuning. We use stable diffusion inpainting as our primary editing model and include results with InstructPix2Pix~\citep{brooks2023instructpix2pix} to show model-agnostic performance.

\section{Methodology}
\label{sec:methodology}

\subsection{Preliminaries}

\paragraph{Image immunization}

Adversarial attacks exploit the vulnerabilities of machine learning models by introducing small, imperceptible perturbations to input data, causing the model to produce incorrect or unintended outputs~\citep{szegedy2014intriguingpropertiesneuralnetworks,Biggio_2013}. In the context of diffusion models, such perturbations can be crafted to disrupt the editing process, ensuring that attempts to modify an adversarially perturbed image fail to achieve intended outcomes. Given an image $\image$, the goal is to transform it into an adversarially immunized version, $\immunizedimage$, by introducing a perturbation $\immunizationnoise$:
\begin{equation}
    \immunizedimage = \image + \immunizationnoise, \quad \text{subject to:} \quad \| \immunizationnoise \|_p < \kappa,
\end{equation}
where $\kappa$ is the perturbation budget that constrains the norm of the perturbation to ensure that it remains imperceptible. The norm $p$ could be chosen as $1$, $2$, or ${\infty}$, depending on the application.

\paragraph{Latent diffusion models}
\acp{ldm}~\citep{rombach2022high} perform the generative process in a lower-dimensional latent space rather than pixel space, achieving computational efficiency while maintaining high-quality outputs. This design is ideal for large-scale tasks like image editing and inpainting.
Training an \ac{ldm} starts by encoding the input image $\image_0$ into a latent representation $\latent_0 = \encoder{\image_0}$ using encoder $\encoder{\cdot}$. The diffusion process operates in this latent space, adding noise over $T$ steps to generate a sequence $\latent_1, \ldots, \latent_T$, with 
\(
\latent_{t+1} = \sqrt{1 - \beta_t} \, \latent_t + \sqrt{\beta_t} \, \epsilon_t, \ \epsilon_t \sim \mathcal{N}(\mathbf{0}, \mathbf{I}),
\)
where $\beta_t$ is the noise schedule at step $t$. The training aims to learn a denoising network $\denoiser$ that predicts the added noise $\epsilon_t$ by minimizing
\(
\mathcal{L}(\theta) = \mathbb{E}_{t, \latent_0, \epsilon \sim \mathcal{N}(\mathbf{0}, \mathbf{I})} \left[ \lVert \epsilon - \denoiser(\latent_t, t) \rVert_2^2 \right].
\)
In the reverse process, a noisy latent vector $\latent_T \sim \mathcal{N}(\mathbf{0}, \mathbf{I})$ is iteratively denoised via the trained denoising network to recover $\latent_0$, which is decoded into the final image $\tilde{\image} = \mathcal{D}(\latent_0)$ with decoder $\mathcal{D(\cdot)}$.

\subsection{Problem Formulation}

\begin{figure*}[t]
\centering
\includegraphics[width=\textwidth]{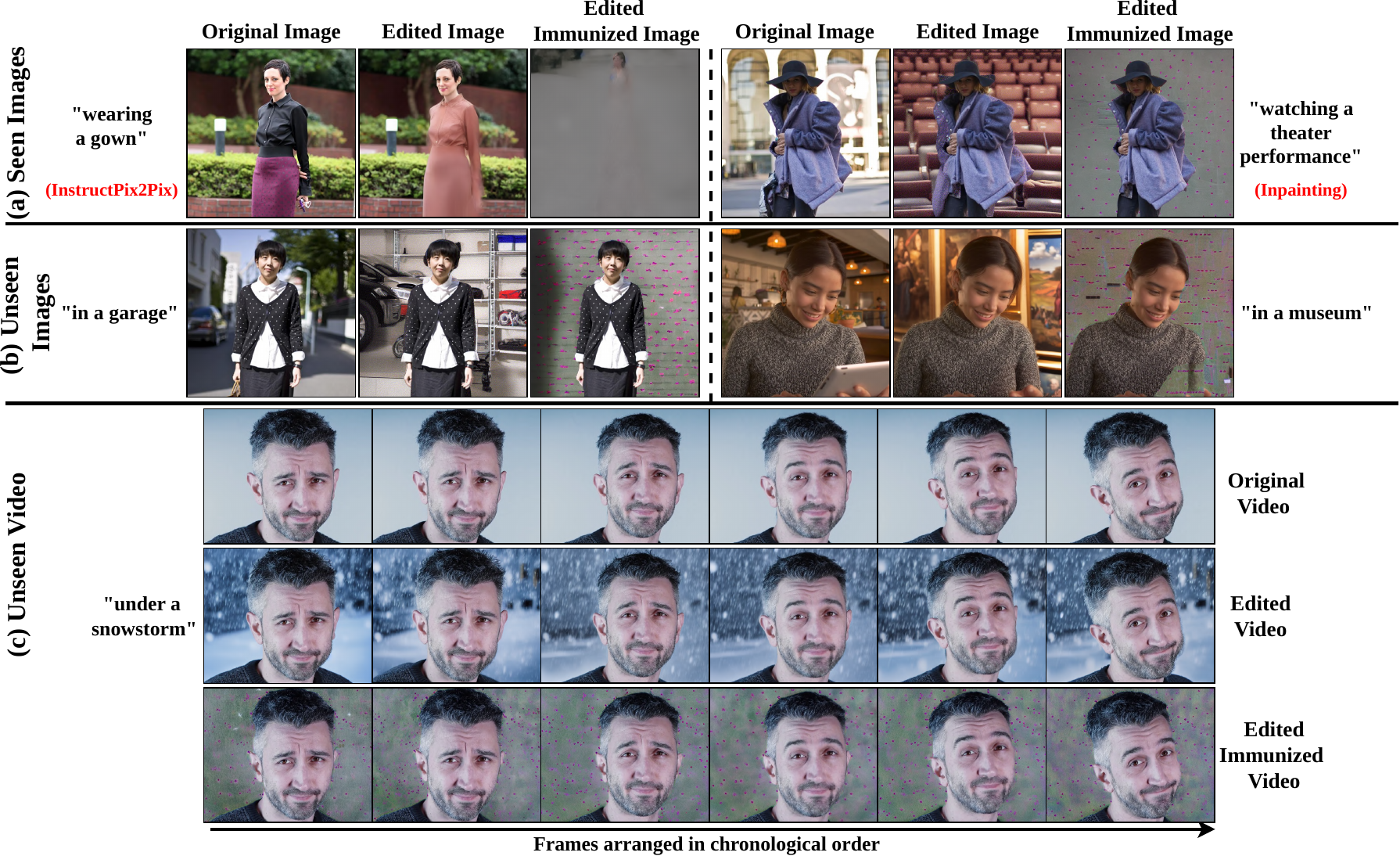}
    \caption{\textit{\textbf{Qualitative results with \methodname.}} Our method effectively immunizes (a) seen images and generalizes to (b) unseen images with diverse text prompts. Additionally, it extends to (c) unseen human videos, demonstrating its adaptability to new content. Furthermore, it supports various poses and perspectives, from full-body shots (a) to close-up face shots (c).}
    \label{fig:qual}
\end{figure*}

Let \(\image \in \mathbb{R}^{H \times W \times C}\) represent an image with height \(H\), width \(W\), and \(C\) color channels. A malicious user using a diffusion-based editing tool, \(\diffusion{\cdot}\), attempts to maliciously edit the image based on a prompt \(\prompt\) and a binary mask \(\mask \in \{0,1\}^{H \times W \times C}\), which defines the target area for editing, with a value of 1 indicating the region of interest and 0 denotes the background or irrelevant areas. Ideally, this target region can represent any meaningful part of the image, such as a human body or a face.
Our objective is to immunize the original (target) image $\image$ by carefully producing a noise $\immunizationnoise$ that satisfies two key criteria: (a) $\immunizationnoise$ remains imperceptible to the user, and (b) the edited immunized image $\editedimage$ fails to accurately reflect the prompt $\prompt$ applied by the malicious users. In other words, the immunized image disrupts the editing model $\diffusion{\cdot}$ such that any attempt to edit the image results in unsuccessful or unintended modifications.
While our approach is broadly applicable to any diffusion-based editing tool, such as inpainting models and InstructPix2Pix~\citep{brooks2023instructpix2pix}, this study follows previous work~\citep{photoguard, Lo_2024_CVPR} by using inpainting as the primary editing tool for problem formulation and quantitative experiments. We focus on scenarios where the sensitive regions such as human body or face remains constant, with other areas considered editable, reflecting real-world malicious editing scenarios. Additional results for other objects and tools (e.g.  InstructPix2Pix) are provided in Fig.~\ref{fig:teaser}, Fig.~\ref{fig:qual}, and in our Appendix~\ref{sec:qualitative_results}.

\subsection{Our Approach}


\begin{figure*}[t]
\centering
\includegraphics[width=1\textwidth]{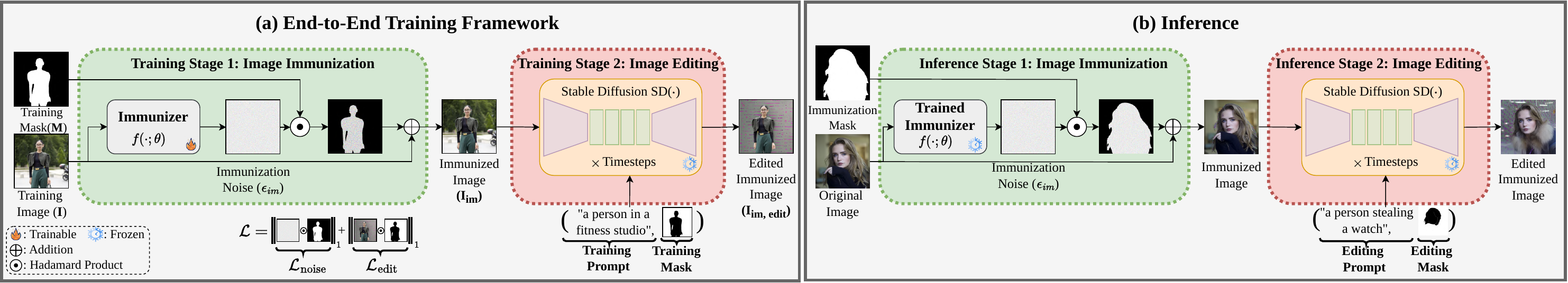} 
\caption{\textbf{\textit{Overview of \methodname.}} 
Our end-to-end training framework is illustrated in (a). The training process consists of two stages. In Stage 1, immunization is applied to the training image \(\image\). In Stage 2, the immunized image \(\immunizedimage\) is edited using a stable diffusion model \(\diffusion{\cdot}\) with the specified text prompt and mask, during which the \(\noiseloss\) and \(\editloss\) are computed. During inference (b), the trained immunizer model generates immunization noise (see Inference Stage 1 in (b)) applied to the original (target) image using an immunization mask. When a malicious user attempts to attack these immunized images with an editing mask, the editing tool (see Inference Stage 2 in (b)) is unable to produce the intended edited content.}
    \label{fig:methodology}
\end{figure*}

\paragraph{End-to-end training framework}
To overcome the speed limitations of previous methods, which require solving an optimization problem independently for each image, we propose an end-to-end training framework. This framework enables an immunizer model \(\unet{\cdot}\) to instantly generate immunization noise for a given input image. Our training algorithm (see Appendix~\ref{a:imp-details}, and Fig.~\ref{fig:methodology} (a)) consists of two stages. In the first stage, we employ a UNet++~\citep{Zhou2018UNetAN} architecture for the ``immunizer'' model \(\unet{\cdot}\), which takes an input image \(\image\) and generates the corresponding immunization noise \(\immunizationnoise\). Subsequently, \(\immunizationnoise\) is multiplied by the immunization mask \(\mask\), which targets the region of interest (e.g.  a person's face). The resulting masked noise is then added to the training image to produce the immunized image, computed as \(\immunizedimage = \image + \immunizationnoise \odot \mask\). Finally, the image is clamped to the \([0, 1]\) range. To ensure the noise remains imperceptible to the human eye, we introduce the following loss:
\begin{equation}
    \noiseloss = \frac{1}{\text{sum}(\mask)}\|(\immunizedimage - \image) \odot \mask\|_p 
    \label{eq:noiseloss}
\end{equation}
where $p$ is empirically chosen to be 1. \(\noiseloss\) penalizes deviations within the masked region, ensuring that the change between the immunized image and the training image is imperceptible. In the second stage, after generating the immunized image \(\immunizedimage\), we apply diffusion-based editing using the editing tool \(\diffusion{\cdot}\). This model takes the immunized image \(\immunizedimage\), the training mask \(\mask\), and the training prompt \(\prompt\) as input, performing edits in the regions specified by the mask. To ensure that the edited image is effectively distorted, we define the loss function:
\begin{equation} 
\editloss = \frac{1}{\text{sum}(\sim \mask)}\|\diffusion{\immunizedimage, \sim\mask, \prompt} \odot (\sim \mask)\|_1, 
\end{equation} 
where $\sim\mask$ represents the complement of the masked area and $\diffusion{\cdot}$ is the stable diffusion inpainting model that modifies the region $\sim \mask$ in $\immunizedimage$ according to the prompt $\prompt$. This loss function is the key to our method, as it ensures that the immunization noise disrupts the editing process by forcing the unmasked regions to be filled with 0s. Note that for editing models that do not rely on masks, we exclude masks from the loss calculations.

To enable training, we curate a dataset of image, mask, and prompt tuples, represented as $\dataset=\{(\image^k, \mask^k, \prompt^k)\}_{k=1}^{N}$. Specifically, we collect 1000 images of individuals from the CCP~\citep{yang2014clothing}
 dataset and use the \acf{sam}~\citep{kirillov2023sam} to generate masks corresponding to the foreground objects in these images. To ensure diverse text descriptions for the editing tasks, we utilize ChatGPT~\cite{chatgpt} (see Appendix~\ref{a:dataset-setup}). At each training step, a sample is selected from the dataset and initially processed by the immunizer model $\unet{\cdot}$ to generate immunization noise $\immunizationnoise^n$, which is added to the masked region of the training image and then clamped. The resulting immunized image $\immunizedimage^n$ is then passed through the editing model $\diffusion{\cdot}$ to produce the edited immunized image $\editedimage^n$. The final loss function, \(\totalloss = \alpha \cdot \noiseloss + \editloss\), is used for backpropagation with respect to the immunizer model's parameters. Backpropagating through the stable diffusion stages allows the immunizer to learn the interaction between the perturbation and the generated pixels. Through this iterative process, the immunizer model learns to generate perturbations that disrupt the editing model. Following the insights from PhotoGuard’s encoder attack, we do not condition the immunizer model on text prompts, as the noise is empirically shown to be prompt-agnostic (see Appendix~\ref{sec:prompt_agonstic_noise}).

\paragraph{Inference}
During inference, the trained immunizer model generates immunization noise for any original (target) image using the mask of the region intended for protection. This noise is then applied to create the immunized image, with the noise restricted to the masked region. The resulting immunized image can be safely shared publicly. When a malicious user inputs this immunized image along with an editing mask into a diffusion-based editing tool (the same tool used during training), the immunization noise disrupts the edited output (see Fig.~\ref{fig:methodology} (b)).
Unlike previous approaches that require the same mask to be used during both training and inference, our method decouples these phases. This separation allows the immunizer model to generalize to unseen content, addressing the limitation of previous methods where malicious users could exploit different masks during editing (e.g.  using an immunization mask of full-body but applying an editing mask of face).

\section{Experimentation}
\label{sec:experiments}



\paragraph{Baselines}
We compare \methodname\ with several existing image immunization methods. As a naive baseline, we include \textbf{Random Noise}, which applies arbitrary noise to images. We also evaluate two variants of PhotoGuard~\citep{photoguard}: \textbf{PhotoGuard-E}, which embeds adversarial perturbations in the latent encoder, and \textbf{PhotoGuard-D}, which disrupts the entire generative process. Additionally, we compare against \textbf{DiffusionGuard}~\citep{choi2025diffusionguard}, an extension of PhotoGuard that augments masks during optimization.
To evaluate robustness against counter-attacks, we develop three additional baselines where editing is applied after immunization: (i) passing the image through a \ac{cnn}-based denoiser~\citep{li2023ntire_dn50}, denoted as \methodname\ w/ D.; (ii) compressing the image as JPEG~\citep{segura2023jpeg} with a 0.75 compression ratio, denoted as \methodname\ w/ JPEG; and (iii) applying the IMPRESS defense~\citep{cao2023impress}, denoted as \methodname\ w/ IMPRESS.

\paragraph{Evaluation metrics and dataset} We focus on four key aspects in evaluation: (a) \textit{the amount of editing failure}, where we follow previous approaches~\citep{photoguard} and utilize SSIM~\citep{wang2004ssim}, PSNR and FSIM~\citep{zhang2011fsim} metrics to measure the visual differences between the edited immunized image and the edited original image; (b) \textit{imperceptibility}, where the amount of the immunization noise quantified by measuring the SSIM between the original image and the immunized image, denoted as SSIM (Noise); (c) \textit{the degree of textual misalignment} evaluated using CLIP~\citep{radford2021clip} by measuring the average similarity between the edited immunized image and the text prompt, denoted as CLIP-T; and (d) \textit{scalability} by reporting the average runtime and GPU memory required to immunize a single image on average from the dataset. We curate a dataset of 875 human images from the CCP~\citep{yang2014clothing} dataset. Of these, 800 images are used for training (including the 75 seen images in our experiments), and 75 unseen images are reserved for testing.

\begin{figure*}
\centering
\includegraphics[width=\textwidth]{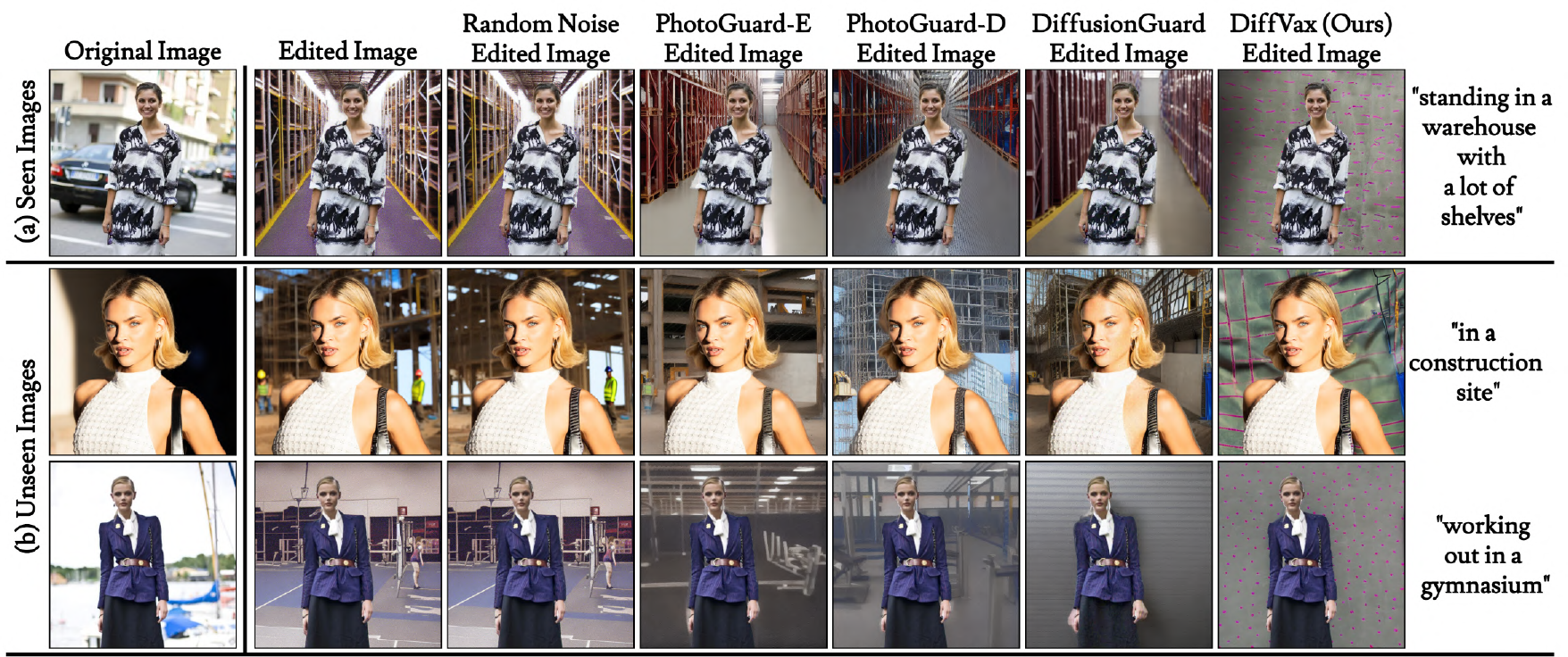} 
    \caption{\textit{\textbf{Qualitative comparison of edited images across immunization methods.}} This figure shows the results of different immunization methods: Random Noise, PhotoGuard-E, PhotoGuard-D, DiffusionGuard, and our proposed method, \methodname. Results for (a) seen and (b) unseen images are shown, with different prompts applied to each (right side). The first column contains the original images, while subsequent columns show the edited outputs under different settings, as depicted on the top. Note that \methodname\ is \emph{substantially more effective} than PhotoGuard-E, -D and DiffusionGuard in degrading the edit.}
    \label{fig:qual-comparison}
    \vspace{-2pt}
\end{figure*}

\paragraph{Qualitative results} 
Figures~\ref{fig:teaser} and ~\ref{fig:qual} illustrate the qualitative success of our method. \methodname\ effectively immunizes images against various editing techniques, including standard inpainting and instruction-based models like InstructPix2Pix~\citep{brooks2023instructpix2pix} (Figure~\ref{fig:teaser}). As further detailed in Appendix~\ref{sec:qualitative_results}, the model demonstrates a strong ability to generalize to unseen images and a wide range of prompts, accommodating various human perspectives from full-body to close-up shots (Figure~\ref{fig:qual}). Although trained primarily on human subjects, our model also extends its robustness to non-human objects. When compared to baseline methods (Figure~\ref{fig:qual-comparison}), our approach is qualitatively superior on both seen and unseen images, generating backgrounds that deviate more significantly from the intended edits. Notably, in many cases with our approach, it is impossible to infer the original prompt from the immunized image's background, a stark contrast to PhotoGuard, which often retains discernible hints of the prompt. More examples, including comparisons and results with other editing models, are provided in Appendix~\ref{sec:qualitative_results}.




\paragraph{DiffVax is more effective in corrupting edits}
As shown in Table~\ref{tab:results}, \methodname\ achieves the lowest SSIM, PSNR, and FSIM values overall, securing second place in the SSIM metric for unseen data, with a small margin behind PG-D, indicating that malicious edits on immunized images are significantly distorted, even on previously unseen data, whereas baseline methods, which require optimization to be re-run for each image, do not differentiate between seen and unseen data. Additionally, CLIP-T results, which measure textual misalignment, further verify these findings by measuring the misalignment semantically in the edited immunized images. \methodname\space outperforms the baselines by maintaining the highest SSIM (Noise) values for both seen and unseen data, highlighting its effectiveness in corrupting malicious edits while keeping the immunized image imperceptible. This superior imperceptibility is achieved because our model learns to generate visually subtle, low-frequency perturbations, in contrast to the scattered, high-frequency noise produced by prior methods (see Appendix~\ref{a:impercep} for a detailed discussion). Thus, \textit{training an immunizer model enables it to learn how to strategically place immunization noise to effectively disrupt diffusion-based editing, by aggregating over the training set.} In contrast, prior optimization-based works only see a single target image at a time.

\begin{table*}[t]
\centering
\caption{\small{\textit{\textbf{Performance comparisons on images.}} The SSIM, PSNR, FSIM, SSIM (Noise), and CLIP-T metrics are reported separately for the \textit{seen} and \textit{unseen} splits of the test dataset. Runtime and GPU requirements are measured as the average time (in seconds) and memory usage (in MiB) needed to immunize a single image. ``N/A" indicates that the corresponding value is unavailable. The symbols $\uparrow$ and $\downarrow$ indicate the direction toward better performance for each metric, respectively. \textbf{Bold} values indicate the best scores, while \underline{underlined} values denote the second-best scores.}}
\resizebox{\textwidth}{!}{%
\begin{tabular}{l|cccccc|cc|cc|c|c}
\toprule
 \multicolumn{1}{c|}{}& \multicolumn{6}{c|}{\textbf{Amount of Editing Failure}}   & \multicolumn{2}{c|}{\textbf{Imperceptibility}} & \multicolumn{2}{c|}{\textbf{Text Misalignment}} &  \multicolumn{2}{c}{\textbf{Scalability}} \\ \cmidrule{2-13}
 
 \multicolumn{1}{c|}{\textbf{Immunization Method}}& \multicolumn{2}{c}{SSIM $\downarrow$} & \multicolumn{2}{c}{PSNR $\downarrow$}  & \multicolumn{2}{c|}{FSIM $\downarrow$} & \multicolumn{2}{c|}{SSIM (Noise) $\uparrow$} & \multicolumn{2}{c|}{CLIP-T $\downarrow$}  & \multicolumn{1}{c|}{Runtime (s) $\downarrow$} & \multicolumn{1}{c}{GPU Req. (MiB) $\downarrow$} \\ 

 &  \textit{seen} & \textit{unseen} & \textit{seen} & \textit{unseen} & \textit{seen} & \textit{unseen} & \textit{seen} & \textit{unseen} & \textit{seen} & \textit{unseen} & (Immunization) & (Immunization) \\
\midrule

Random Noise & 0.586 & 0.585 & 16.09 & 16.40 & 0.460 & 0.458 & 0.902 & 0.903 & 31.68 & 31.62 & N/A & N/A  \\
PhotoGuard-E   & 0.558 & 0.565 & 15.29 & 15.63 & 0.413 & 0.408 & 0.956 & 0.956 & 31.69 & 30.88 &  207.00 &  9,548  \\ 
PhotoGuard-D   & \underline{\textit{0.531}} & \textbf{0.523} & 14.70 & 14.92 & \underline{\textit{0.386}} & \underline{\textit{0.379}} & \underline{\textit{0.978}} & \underline{\textit{0.979}} & 29.61 & 29.27 & 911.60 & 15,114 \\
DiffusionGuard   & 0.551 & 0.556 & \underline{\textit{14.37}} & \underline{\textit{14.71}} & 0.389 & 0.386 & 0.965 & 0.966 & \underline{\textit{26.98}} & \underline{\textit{27.10}} & \underline{\textit{131.10}} & \underline{\textit{6,750}} \\
\cellcolor{green!25}\methodname\ (Ours) & \cellcolor{green!25} \textbf{0.510} & 
\cellcolor{green!25} \underline{\textit{0.526}} & \cellcolor{green!25} \textbf{13.96} & \cellcolor{green!25} \textbf{14.32} & \cellcolor{green!25} \textbf{0.353} & \cellcolor{green!25} \textbf{0.362} & \cellcolor{green!25} \textbf{0.989} &  \cellcolor{green!25} \textbf{0.989} & \cellcolor{green!25} \textbf{23.13} & \cellcolor{green!25} \textbf{24.17} & \cellcolor{green!25} \textbf{0.07} & \cellcolor{green!25} \textbf{5,648} \\
\bottomrule
\end{tabular}}
\vspace{4pt}
\label{tab:results}
\end{table*}

\paragraph{DiffVax is more scalable}
In addition to its strong qualitative performance, \methodname\ offers significant advantages in speed and memory efficiency. It completes the immunization process in just 0.07 seconds per image on average, compared to 207.0 seconds for PhotoGuard-E, 911.6 seconds for PhotoGuard-D, and 131.1 seconds for DiffusionGuard. In terms of GPU memory usage, \methodname\ requires only 5,648 MiB, much lower than PhotoGuard-E (9,548 MiB), PhotoGuard-D (15,114 MiB), and DiffusionGuard (6,750 MiB). This makes \textit{\methodname\ a practical and scalable solution for large-scale applications.}

\begin{table*}[t]
\centering
\caption{\small{\textit{\textbf{Performance comparisons on edits with counter-attacks.}} We report the SSIM, SSIM (Noise) and CLIP-T metrics for the denoiser (D.), JPEG (compression ratio of 0.75) counter-attacks separately for the \textit{seen} and \textit{unseen} splits of the test dataset.}}
\resizebox{\columnwidth}{!}{%
\begin{tabular}{l|cc|cc|cc|cc|cc}
\toprule
 
 \multicolumn{1}{c|}{\textbf{Method}} &  \multicolumn{2}{c|}{SSIM $\downarrow$} &
 \multicolumn{2}{c|}{PSNR $\downarrow$} & \multicolumn{2}{c|}{FSIM $\downarrow$} & \multicolumn{2}{c|}{SSIM (Noise) $\uparrow$} & \multicolumn{2}{c}{CLIP-T $\downarrow$}  \\ 

 &  \textit{seen} & \textit{unseen} & \textit{seen} & \textit{unseen} & \textit{seen} & \textit{unseen} & \textit{seen} & \textit{unseen} & \textit{seen} & \textit{unseen} \\
\midrule

\cellcolor{gray!15} PG-D w/ D.   & \cellcolor{gray!15} 0.702 & \cellcolor{gray!15} 0.709 & \cellcolor{gray!15} 18.27 & \cellcolor{gray!15} 18.43 & \cellcolor{gray!15} 0.528 & \cellcolor{gray!15} 0.528 & \cellcolor{gray!15} \textbf{0.966} & \cellcolor{gray!15} \textbf{0.965} & \cellcolor{gray!15} 31.48 & \cellcolor{gray!15} 31.20 \\ 
\cellcolor{gray!35}  DiffusionGuard w/ D. & \cellcolor{gray!35} 0.708 & \cellcolor{gray!35} 0.719 & \cellcolor{gray!35} 18.26 & \cellcolor{gray!35} 18.69 & \cellcolor{gray!35} 0.530 & \cellcolor{gray!35} 0.531 & \cellcolor{gray!35} 0.964 & \cellcolor{gray!35} 0.964 & \cellcolor{gray!35} 31.08 & \cellcolor{gray!35} 30.99 \\ 
\cellcolor{gray!15}  \methodname \space w/ D. & \cellcolor{gray!15} \textbf{0.552} & \cellcolor{gray!15} \textbf{0.565} & \cellcolor{gray!15} \textbf{14.48} & \cellcolor{gray!15} \textbf{14.91} & \cellcolor{gray!15} \textbf{0.388} & \cellcolor{gray!15} \textbf{0.392} & \cellcolor{gray!15} 0.960 & \cellcolor{gray!15} 0.960 & \cellcolor{gray!15} \textbf{27.32} & \cellcolor{gray!15} \textbf{27.74}\\ \midrule
\cellcolor{gray!35} PG-D w/ JPEG & \cellcolor{gray!35} 0.664 & \cellcolor{gray!35} 0.674 & \cellcolor{gray!35} 17.32 & \cellcolor{gray!35} 17.68 & \cellcolor{gray!35} 0.495 & \cellcolor{gray!35} 0.501 \cellcolor{gray!35} & \cellcolor{gray!35} 0.956 & \cellcolor{gray!35} 0.956 &  \cellcolor{gray!35} 32.15 & \cellcolor{gray!35} 32.48 \\
\cellcolor{gray!15} DiffusionGuard w/ JPEG & \cellcolor{gray!15} 0.680 & \cellcolor{gray!15} 0.684 & \cellcolor{gray!15} 17.45 & \cellcolor{gray!15} 17.83 & \cellcolor{gray!15} 0.505 & \cellcolor{gray!15} 0.503 & \cellcolor{gray!15} 0.951 & \cellcolor{gray!15} 0.951 &  \cellcolor{gray!15} 31.52 & \cellcolor{gray!15} 31.53 \\
\cellcolor{gray!35} \methodname \space w/ JPEG & \cellcolor{gray!35}  \textbf{0.522} & \cellcolor{gray!35} \textbf{0.538} & \cellcolor{gray!35} \textbf{14.17} & \cellcolor{gray!35} \textbf{14.61} & \cellcolor{gray!35} \textbf{0.374} & \cellcolor{gray!35} \textbf{0.382} &\cellcolor{gray!35} \textbf{0.959} & \cellcolor{gray!35} \textbf{0.959} & \cellcolor{gray!35} \textbf{26.04} & \cellcolor{gray!35} \textbf{26.05} \\
\midrule
\cellcolor{gray!15} PG-D w/ IMPRESS & \cellcolor{gray!15} 0.578 & \cellcolor{gray!15} 0.563 & \cellcolor{gray!15} 15.89 & \cellcolor{gray!15} 16.07 & \cellcolor{gray!15} 0.436 & \cellcolor{gray!15} 0.426 & \cellcolor{gray!15} 0.640 & \cellcolor{gray!15} 0.634 &  \cellcolor{gray!15} 31.35 & \cellcolor{gray!15} 31.26 \\
\cellcolor{gray!35} DiffusionGuard \space w/ IMPRESS & \cellcolor{gray!35}  0.604 & \cellcolor{gray!35} 0.595 & \cellcolor{gray!35} 15.89 & \cellcolor{gray!35} 16.09 & \cellcolor{gray!35} 0.453 & \cellcolor{gray!35} 0.442 & \cellcolor{gray!35} 0.636 & \cellcolor{gray!35} 0.630 & \cellcolor{gray!35} 30.88 & \cellcolor{gray!35} 30.50 \\
\cellcolor{gray!15} \methodname \space w/ IMPRESS & \cellcolor{gray!15}  \textbf{0.488} & \cellcolor{gray!15} \textbf{0.500} & \cellcolor{gray!15} \textbf{14.04} & \cellcolor{gray!15} \textbf{14.38} & \cellcolor{gray!15} \textbf{0.355} & \cellcolor{gray!15} \textbf{0.359} & \cellcolor{gray!15} \textbf{0.644} & \cellcolor{gray!15} \textbf{0.637} & \cellcolor{gray!15} \textbf{24.88} & \cellcolor{gray!15} \textbf{25.27} \\
\bottomrule
\end{tabular}
}
\vspace{-3pt}
\label{tab:results-attack}
\end{table*}

\paragraph{DiffVax is more robust to counter-attacks}
Table~\ref{tab:results-attack} shows that \methodname\ is robust to common counter-attacks, including CNN-based denoising, JPEG compression, and IMPRESS~\citep{cao2023impress}. \methodname\ consistently outperforms PhotoGuard-D across all scenarios, as further evidenced by the detailed results in Appendix~\ref{sec:robustness_counterattacks}. This robustness arises from \methodname’s ability to learn spatially targeted, low-frequency perturbations. Unlike existing approaches that produce more uniform, high-frequency noise, our method's perturbations are less susceptible to removal by techniques like JPEG compression, which discards high-frequency content, or by denoisers trained to suppress uniform noise. Crucially, as shown in Appendix~\ref{a:pert-budget}, \methodname\ achieves superior edit disruption with a much smaller mean magnitude of noise than baselines with larger fixed budgets. This highlights that its strength lies in the strategic placement of noise, not simply its magnitude, supporting our claim that \methodname\ learns a more efficient and targeted noise distribution. Furthermore, our extensive robustness evaluations in Appendix~\ref{a:robustness} show that \methodname\ also maintains its effectiveness against attackers who vary their inference-time settings, consistently outperforming baselines across different sampling steps and diffusion samplers.

\paragraph{User study results} We also conduct a user study with 67 participants on~\cite{Prolific}, in which participants compare the ``unrealisticness'' level of baselines, and the edited image across 20 randomly selected image pairs, including both seen and unseen samples. For each model, we report the average rank, with our model achieving the top position with an average rank of 1.64, demonstrating clear superiority (see Appendix~\ref{sec:user_study}), followed by PhotoGuard-D with a rank of 2.63.

\paragraph{Ablation study} To assess the contribution of each component in our framework, we conduct an ablation study by individually removing $\mathcal{L}_\text{edit}$ and $\mathcal{L}_\text{noise}$. As shown in Table~\ref{tab:ablation}, when $\noiseloss$ is removed, the model achieves slightly better performance on unseen data in terms of failed immunized editing (measured by SSIM, PSNR, FSIM and CLIP-T). However, the immunization noise is no longer imperceptible, as indicated by the change in the SSIM (Noise) metric. Conversely, when $\editloss$ is removed, the SSIM (Noise) metric reaches its highest value, indicating minimal noise, but the model fails to prevent malicious editing, as reflected in the SSIM, PSNR, FSIM and CLIP-T metrics. Thus, \textit{combining both terms in the final loss function is crucial for balancing imperceptibility and robustness in the training process} (see Appendix~\ref{sec:complementary_ablation_study}).

\begin{table*}[t]
\centering
\caption{\small{\textit{\textbf{Ablation study.}} We report the SSIM and SSIM (Noise) metrics for each loss term ablation, with results presented individually for the seen and unseen splits of the dataset.}}
\resizebox{\columnwidth}{!}{%
\begin{tabular}{lcccccccccc}
\toprule
 \multicolumn{1}{c}{\textbf{Method}}& \multicolumn{2}{c}{SSIM $\downarrow$} & \multicolumn{2}{c}{PSNR $\downarrow$} & \multicolumn{2}{c}{FSIM $\downarrow$} & \multicolumn{2}{c}{SSIM (Noise) $\uparrow$} & \multicolumn{2}{c}{CLIP-T $\downarrow$}  \\ 

 &  \textit{s} & \textit{u} & \textit{s} & \textit{u}  &  \textit{s} & \textit{u} & \textit{s} & \textit{u} & \textit{s} & \textit{u}    \\
\midrule

 \methodname \space w/o $\mathcal{L}_\text{noise}$ & \textbf{0.508} & \textbf{0.520} & \textbf{13.57} & \textbf{13.82} & \textbf{0.335} & \textbf{0.344} & 0.785 & 0.786 & 24.34 & 25.78 \\
 \methodname \space w/o $\mathcal{L}_\text{edit}$ & 0.944 & 0.932 & 31.36 & 31.05 & 0.821 & 0.806 & \textbf{0.999} & \textbf{0.999} & 32.01 & 32.27 \\ \midrule
\methodname & 0.510 & 0.526 & 13.96 & 14.32 & 0.353 & 0.362 & 0.989 &  0.989 & \textbf{23.13} & \textbf{24.17}  \\
\bottomrule
\end{tabular}%
}
\vspace{-1pt}
\label{tab:ablation}
\end{table*}

\begin{figure*}
\centering
\includegraphics[width=\textwidth]{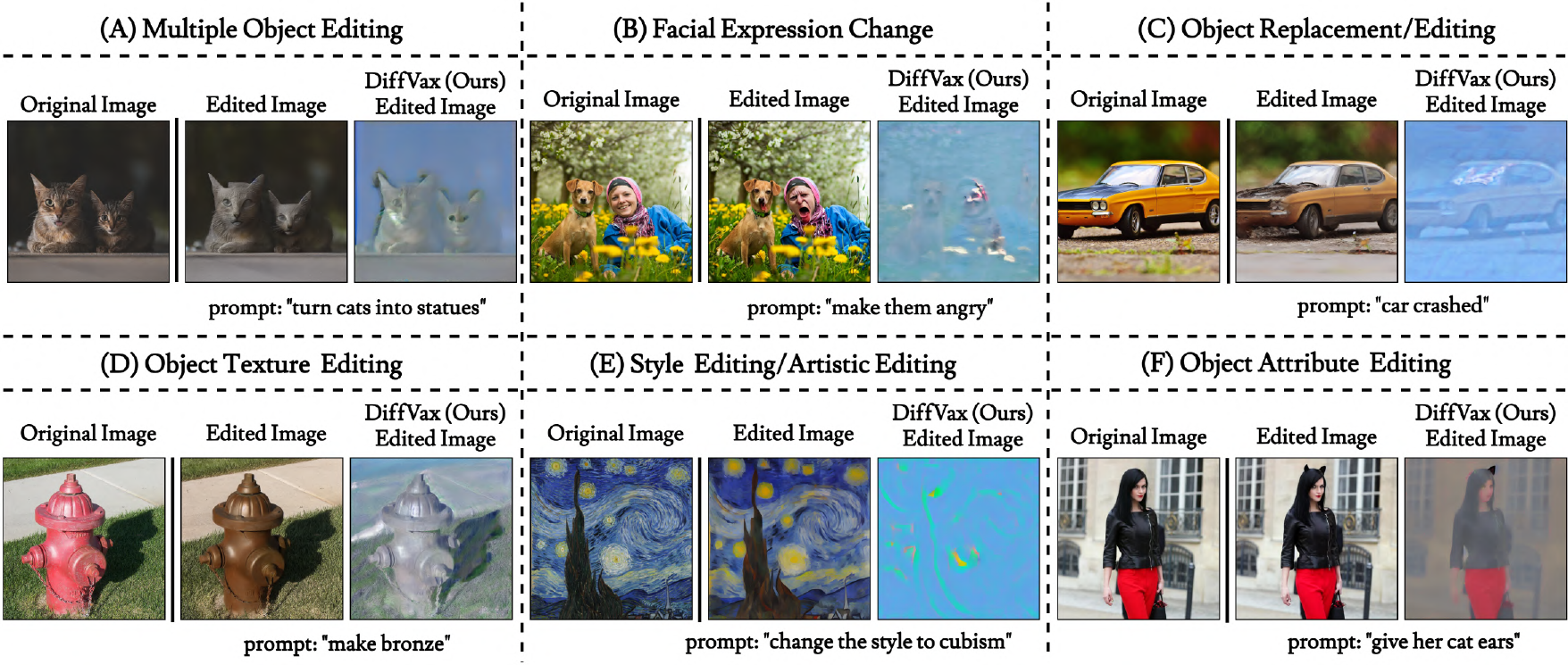}
\caption{\textit{\textbf{DiffVax performance across diverse editing applications.}} Our method effectively immunizes images against various manipulation types, including (A) multiple object editing, (B) facial expression changes, (C) object replacement, (D) texture editing, (E) artistic style transfer, and (F) attribute editing.}
\label{fig:add-applications}
\vspace{-2pt}
\end{figure*}

\section{Conclusion and Discussion}
\label{sec:conclusion}

\paragraph{Discussion on generalization} While a universal immunizer remains an open challenge, \methodname\ demonstrates superior generalization over prior optimization-based work across three key dimensions. First, it addresses \textbf{generalization to unseen models}. Universal cross-model transferability is a difficult open problem, and like prior work, \methodname\ is primarily model-specific and does not perfectly generalize to all unseen models. However, as detailed in Appendix~\ref{a:unseen-models}, it demonstrates significantly better performance in the challenging black-box transfer task from a model trained on Stable Diffusion (SD) v1.5 to an unseen SD v2 model. In this scenario, our learned immunization successfully transfers its protective effect, whereas optimization-based methods like PhotoGuard and DiffusionGuard fail completely, showing a clear improvement in cross-model robustness. Second, its feed-forward nature enables \textbf{generalization to unseen content}, a significant advantage over methods requiring costly per-image optimization. As discussed in Appendix~\ref{a:unseen-content}, the success of \methodname\ proves that the set of effective perturbations has a learnable structure, allowing it to immunize new images, prompts, and even videos with a single pass. Finally, \methodname\ is uniquely robust in its \textbf{generalization to unseen masks}. Unlike prior work, it does not overfit to the training mask's shape or scale, maintaining its edit-disrupting effectiveness even when test-time editing masks differ significantly from the immunization mask, as shown in Appendix~\ref{a:unseen-masking}.

\paragraph{Discussion on applications} To further validate the versatility of \methodname, we evaluate its effectiveness across a broad spectrum of real-world editing scenarios, as illustrated in Figure~\ref{fig:add-applications}. Our experiments demonstrate robust protection against diverse manipulation types, ranging from \textbf{global transformations} such as artistic style transfer (e.g., converting an image to Cubism) to \textbf{localized edits} like facial expression changes and object attribute modification. Furthermore, the framework proves effective in complex settings involving multiple object editing, object replacement, and texture alteration. This wide applicability confirms that \methodname\ does not merely memorize specific noise patterns for simple inpainting but learns to disrupt the underlying semantic guidance required for various high-level editing tasks.

\paragraph{Discussion on editing models}
Following prior work, our main evaluations are conducted using inpainting-based editing methods. However, we emphasize that our framework is model-agnostic and can be applied to various editing tools. To demonstrate this, we include additional results using the instruction-based model InstructPix2Pix (IP2P)~\citep{brooks2023instructpix2pix} (see Figure~\ref{fig:qualitative_ip2p} in the Appendix) and the training-free model MagicBrush~\citep{zhang2023magicbrush} (see Table~\ref{tab:magicbrush-comparison} in the Appendix).
We find that IP2P is particularly well-suited for complex or localized editing tasks, such as background modifications, stylistic changes, or edits outside sensitive regions, whereas inpainting-based approaches are more specialized for background editing tasks. Specifically, inpainting methods can introduce unintended alterations in sensitive areas like faces when the provided mask only partially covers the target region. This can conflict with the intent of a malicious user, whose goal is often to preserve identity while making selective edits. 

\paragraph{Conclusion} In this work, we present \methodname, an optimization-free image immunization framework that protects against diffusion-based editing. Central to our approach is a trained ``image immunizer'' model that generates imperceptible perturbations to disrupt the editing process. At inference, \methodname\ requires only a single forward pass, enabling scalability to large-scale deployments. Leveraging this efficiency, we extend our framework to video, demonstrating promising results for the first time (see Appendix~\ref{sec:video_eval}). 
Moreover, \methodname\ is compatible with any diffusion-based editing tool and demonstrates strong robustness against counter-attacks. Overall, it establishes a new benchmark for scalable, real-time, and effective content protection.




\subsubsection*{Acknowledgments}
We acknowledge partial support from the Health Care Engineering Systems Center (Grainger College of Engineering, UIUC).


\newpage
\bibliography{main}

@article{Wojewidka2020TheDT,
  title={The deepfake threat to face biometrics},
  author={John Wojewidka},
  journal={Biometric Technology Today},
  year={2020},
  volume={2020},
  pages={5-7},
  url={https://api.semanticscholar.org/CorpusID:212981964}
}

@article{Delfino2022DeepfakesOT,
  title={Deepfakes on Trial: a Call to Expand the Trial Judge’S Gatekeeping Role to Protect Legal Proceedings from Technological Fakery},
  author={Rebecca A. Delfino},
  journal={SSRN Electronic Journal},
  year={2022},
  url={https://api.semanticscholar.org/CorpusID:246806628}
}

@online{KoreanDF,
  author = {Jean Mackenzie, Leehyun Choi},
  title = {Inside the deepfake porn crisis engulfing Korean schools},
  year = 2024,
  howpublished = {\url{https://web.archive.org/web/20240928170449/https://www.bbc.com/news/articles/cpdlpj9zn9go}},
  urldate = {2024-10-07}
}

@online{Deepfaked,
  author = {Jess Davies and Sarah McDermott},
  title = {Deepfaked: ‘They put my face on a porn video’},
  year = {2022},
  howpublished = {\url{https://www.bbc.com/news/uk-62821117}},
  note = {Accessed: 2024-11-14}
}

@online{ViceDF,
  author = {Samantha Cole},
  title = {We Are Truly Fucked: Everyone Is Making AI-Generated Fake Porn Now},
  year = 2018,
  howpublished = {\url{https://web.archive.org/web/20240926135620/https://www.vice.com/en/article/reddit-fake-porn-app-daisy-ridley/}},
  urldate = {2024-10-07},
  note = {Accessed: 2024-11-14}
}

@article{Naitali2023DeepfakeAG,
  title={Deepfake Attacks: Generation, Detection, Datasets, Challenges, and Research Directions},
  author={Amal Naitali and Mohammed Ridouani and Fatima Salahdine and Naima Kaabouch},
  journal={Comput.},
  year={2023},
  volume={12},
  pages={216},
  url={https://api.semanticscholar.org/CorpusID:264478099}
}

@article{Hasan2019CombatingDV,
  title={Combating Deepfake Videos Using Blockchain and Smart Contracts},
  author={Haya R. Hasan and Khaled Salah},
  journal={IEEE Access},
  year={2019},
  volume={7},
  pages={41596-41606},
  url={https://api.semanticscholar.org/CorpusID:88489143}
}

@inproceedings{li2024safegen,
  author       = {Li, Xinfeng and Yang, Yuchen and Deng, Jiangyi and Yan, Chen and Chen, Yanjiao and Ji, Xiaoyu and Xu, Wenyuan},
  title        = {{SafeGen: Mitigating Sexually Explicit Content Generation in Text-to-Image Models}},
  booktitle    = {Proceedings of the 2024 {ACM} {SIGSAC} Conference on Computer and Communications Security (CCS)},
  year         = {2024},
}

@InProceedings{photoguard,
  title = 	 {Raising the Cost of Malicious {AI}-Powered Image Editing},
  author =       {Salman, Hadi and Khaddaj, Alaa and Leclerc, Guillaume and Ilyas, Andrew and Madry, Aleksander},
  booktitle = 	 {Proceedings of the 40th International Conference on Machine Learning},
  pages = 	 {29894--29918},
  year = 	 {2023},
  editor = 	 {Krause, Andreas and Brunskill, Emma and Cho, Kyunghyun and Engelhardt, Barbara and Sabato, Sivan and Scarlett, Jonathan},
  volume = 	 {202},
  series = 	 {Proceedings of Machine Learning Research},
  month = 	 {23--29 Jul},
  publisher =    {PMLR},
  url = 	 {https://proceedings.mlr.press/v202/salman23a.html}
}

@InProceedings{Lo_2024_CVPR,
    author    = {Lo, Ling and Yeo, Cheng Yu and Shuai, Hong-Han and Cheng, Wen-Huang},
    title     = {Distraction is All You Need: Memory-Efficient Image Immunization against Diffusion-Based Image Editing},
    booktitle = {Proceedings of the IEEE/CVF Conference on Computer Vision and Pattern Recognition (CVPR)},
    month     = {June},
    year      = {2024},
    pages     = {24462-24471}
}

@inproceedings{goodfellow2015explainingharnessingadversarialexamples,
  author       = {Ian J. Goodfellow and
                  Jonathon Shlens and
                  Christian Szegedy},
  editor       = {Yoshua Bengio and
                  Yann LeCun},
  title        = {Explaining and Harnessing Adversarial Examples},
  booktitle    = {3rd International Conference on Learning Representations, {ICLR} 2015,
                  San Diego, CA, USA, May 7-9, 2015, Conference Track Proceedings},
  year         = {2015},
  url          = {http://arxiv.org/abs/1412.6572},
  bibsource    = {dblp computer science bibliography, https://dblp.org}
}

@inproceedings{madry2019deeplearningmodelsresistant,
  author       = {Aleksander Madry and
                  Aleksandar Makelov and
                  Ludwig Schmidt and
                  Dimitris Tsipras and
                  Adrian Vladu},
  title        = {Towards Deep Learning Models Resistant to Adversarial Attacks},
  booktitle    = {6th International Conference on Learning Representations, {ICLR} 2018,
                  Vancouver, BC, Canada, April 30 - May 3, 2018, Conference Track Proceedings},
  publisher    = {OpenReview.net},
  year         = {2018},
  url          = {https://openreview.net/forum?id=rJzIBfZAb},
  bibsource    = {dblp computer science bibliography, https://dblp.org}
}

@inproceedings{carlini2017evaluatingrobustnessneuralnetworks,
  author       = {Nicholas Carlini and
                  David A. Wagner},
  title        = {Towards Evaluating the Robustness of Neural Networks},
  booktitle    = {2017 {IEEE} Symposium on Security and Privacy, {SP} 2017, San Jose,
                  CA, USA, May 22-26, 2017},
  pages        = {39--57},
  publisher    = {{IEEE} Computer Society},
  year         = {2017},
  url          = {https://doi.org/10.1109/SP.2017.49},
  doi          = {10.1109/SP.2017.49},
  bibsource    = {dblp computer science bibliography, https://dblp.org}
}

@inproceedings{moosavidezfooli2016deepfoolsimpleaccuratemethod,
  author       = {Seyed{-}Mohsen Moosavi{-}Dezfooli and
                  Alhussein Fawzi and
                  Pascal Frossard},
  title        = {DeepFool: {A} Simple and Accurate Method to Fool Deep Neural Networks},
  booktitle    = {2016 {IEEE} Conference on Computer Vision and Pattern Recognition,
                  {CVPR} 2016, Las Vegas, NV, USA, June 27-30, 2016},
  pages        = {2574--2582},
  publisher    = {{IEEE} Computer Society},
  year         = {2016},
  url          = {https://doi.org/10.1109/CVPR.2016.282},
  doi          = {10.1109/CVPR.2016.282},
  bibsource    = {dblp computer science bibliography, https://dblp.org}
}

@inproceedings{xiao2019generatingadversarialexamplesadversarial,
  author       = {Chaowei Xiao and
                  Bo Li and
                  Jun{-}Yan Zhu and
                  Warren He and
                  Mingyan Liu and
                  Dawn Song},
  editor       = {J{\'{e}}r{\^{o}}me Lang},
  title        = {Generating Adversarial Examples with Adversarial Networks},
  booktitle    = {Proceedings of the Twenty-Seventh International Joint Conference on
                  Artificial Intelligence, {IJCAI} 2018, July 13-19, 2018, Stockholm,
                  Sweden},
  pages        = {3905--3911},
  publisher    = {ijcai.org},
  year         = {2018},
  url          = {https://doi.org/10.24963/ijcai.2018/543},
  doi          = {10.24963/IJCAI.2018/543},
  bibsource    = {dblp computer science bibliography, https://dblp.org}
}

@inproceedings{dong2018boostingadversarialattacksmomentum,
  author       = {Yinpeng Dong and
                  Fangzhou Liao and
                  Tianyu Pang and
                  Hang Su and
                  Jun Zhu and
                  Xiaolin Hu and
                  Jianguo Li},
  title        = {Boosting Adversarial Attacks With Momentum},
  booktitle    = {2018 {IEEE} Conference on Computer Vision and Pattern Recognition,
                  {CVPR} 2018, Salt Lake City, UT, USA, June 18-22, 2018},
  pages        = {9185--9193},
  publisher    = {Computer Vision Foundation / {IEEE} Computer Society},
  year         = {2018},
  url          = {http://openaccess.thecvf.com/content\_cvpr\_2018/html/Dong\_Boosting\_Adversarial\_Attacks\_CVPR\_2018\_paper.html},
  doi          = {10.1109/CVPR.2018.00957},
  bibsource    = {dblp computer science bibliography, https://dblp.org}
}

@inproceedings{andriushchenko2020squareattackqueryefficientblackbox,
  author       = {Maksym Andriushchenko and
                  Francesco Croce and
                  Nicolas Flammarion and
                  Matthias Hein},
  editor       = {Andrea Vedaldi and
                  Horst Bischof and
                  Thomas Brox and
                  Jan{-}Michael Frahm},
  title        = {Square Attack: {A} Query-Efficient Black-Box Adversarial Attack via
                  Random Search},
  booktitle    = {Computer Vision - {ECCV} 2020 - 16th European Conference, Glasgow,
                  UK, August 23-28, 2020, Proceedings, Part {XXIII}},
  series       = {Lecture Notes in Computer Science},
  volume       = {12368},
  pages        = {484--501},
  publisher    = {Springer},
  year         = {2020},
  url          = {https://doi.org/10.1007/978-3-030-58592-1\_29},
  doi          = {10.1007/978-3-030-58592-1\_29},
  bibsource    = {dblp computer science bibliography, https://dblp.org}
}

@inproceedings{croce2020reliableevaluationadversarialrobustness,
  author       = {Francesco Croce and
                  Matthias Hein},
  title        = {Reliable evaluation of adversarial robustness with an ensemble of
                  diverse parameter-free attacks},
  booktitle    = {Proceedings of the 37th International Conference on Machine Learning,
                  {ICML} 2020, 13-18 July 2020, Virtual Event},
  series       = {Proceedings of Machine Learning Research},
  volume       = {119},
  pages        = {2206--2216},
  publisher    = {{PMLR}},
  year         = {2020},
  url          = {http://proceedings.mlr.press/v119/croce20b.html},
  bibsource    = {dblp computer science bibliography, https://dblp.org}
}

@inproceedings{moosavidezfooli2017universaladversarialperturbations,
  author       = {Seyed{-}Mohsen Moosavi{-}Dezfooli and
                  Alhussein Fawzi and
                  Omar Fawzi and
                  Pascal Frossard},
  title        = {Universal Adversarial Perturbations},
  booktitle    = {2017 {IEEE} Conference on Computer Vision and Pattern Recognition,
                  {CVPR} 2017, Honolulu, HI, USA, July 21-26, 2017},
  pages        = {86--94},
  publisher    = {{IEEE} Computer Society},
  year         = {2017},
  url          = {https://doi.org/10.1109/CVPR.2017.17},
  doi          = {10.1109/CVPR.2017.17},
  bibsource    = {dblp computer science bibliography, https://dblp.org}
}

@inproceedings{hayes2018learninguniversaladversarialperturbations,
  author       = {Jamie Hayes and
                  George Danezis},
  title        = {Learning Universal Adversarial Perturbations with Generative Models},
  booktitle    = {2018 {IEEE} Security and Privacy Workshops, {SP} Workshops 2018, San
                  Francisco, CA, USA, May 24, 2018},
  pages        = {43--49},
  publisher    = {{IEEE} Computer Society},
  year         = {2018},
  url          = {https://doi.org/10.1109/SPW.2018.00015},
  doi          = {10.1109/SPW.2018.00015},
  bibsource    = {dblp computer science bibliography, https://dblp.org}
}

@inproceedings{yeh2021attackbestdefensenullifying,
  author       = {Chin{-}Yuan Yeh and
                  Hsi{-}Wen Chen and
                  Hong{-}Han Shuai and
                  De{-}Nian Yang and
                  Ming{-}Syan Chen},
  title        = {Attack as the Best Defense: Nullifying Image-to-image Translation
                  GANs via Limit-aware Adversarial Attack},
  booktitle    = {2021 {IEEE/CVF} International Conference on Computer Vision, {ICCV}
                  2021, Montreal, QC, Canada, October 10-17, 2021},
  pages        = {16168--16177},
  publisher    = {{IEEE}},
  year         = {2021},
  url          = {https://doi.org/10.1109/ICCV48922.2021.01588},
  doi          = {10.1109/ICCV48922.2021.01588},
  bibsource    = {dblp computer science bibliography, https://dblp.org}
}

@inproceedings{aneja2022tafimtargetedadversarialattacks,
  author       = {Shivangi Aneja and
                  Lev Markhasin and
                  Matthias Nie{\ss}ner},
  editor       = {Shai Avidan and
                  Gabriel J. Brostow and
                  Moustapha Ciss{\'{e}} and
                  Giovanni Maria Farinella and
                  Tal Hassner},
  title        = {{TAFIM:} Targeted Adversarial Attacks Against Facial Image Manipulations},
  booktitle    = {Computer Vision - {ECCV} 2022 - 17th European Conference, Tel Aviv,
                  Israel, October 23-27, 2022, Proceedings, Part {XIV}},
  series       = {Lecture Notes in Computer Science},
  volume       = {13674},
  pages        = {58--75},
  publisher    = {Springer},
  year         = {2022},
  url          = {https://doi.org/10.1007/978-3-031-19781-9\_4},
  doi          = {10.1007/978-3-031-19781-9\_4},
  bibsource    = {dblp computer science bibliography, https://dblp.org}
}

@inproceedings {shawn2023glaze,
author = {Shawn Shan and Jenna Cryan and Emily Wenger and Haitao Zheng and Rana Hanocka and Ben Y. Zhao},
title = {Glaze: Protecting Artists from Style Mimicry by {Text-to-Image} Models},
booktitle = {32nd USENIX Security Symposium (USENIX Security 23)},
year = {2023},
isbn = {978-1-939133-37-3},
address = {Anaheim, CA},
pages = {2187--2204},
url = {https://www.usenix.org/conference/usenixsecurity23/presentation/shan},
publisher = {USENIX Association},
month = aug
}

@misc{pei2024deepfakegenerationdetectionbenchmark,
      title={Deepfake Generation and Detection: A Benchmark and Survey}, 
      author={Gan Pei and Jiangning Zhang and Menghan Hu and Zhenyu Zhang and Chengjie Wang and Yunsheng Wu and Guangtao Zhai and Jian Yang and Chunhua Shen and Dacheng Tao},
      year={2024},
      eprint={2403.17881},
      archivePrefix={arXiv},
      primaryClass={cs.CV},
      url={https://arxiv.org/abs/2403.17881}, 
}

@article{appel2022detectionofpolitical,
    author = {Appel, Markus and Prietzel, Fabian},
    title = "{The detection of political deepfakes}",
    journal = {Journal of Computer-Mediated Communication},
    volume = {27},
    number = {4},
    pages = {zmac008},
    year = {2022},
    month = {07},
    issn = {1083-6101},
    doi = {10.1093/jcmc/zmac008},
    url = {https://doi.org/10.1093/jcmc/zmac008},
    eprint = {https://academic.oup.com/jcmc/article-pdf/27/4/zmac008/45085688/zmac008.pdf},
}

@inproceedings{Blancaflor2024deepfake,
author = {Blancaflor, Eric and Garcia, Joshua Ivan and Magno, Frances Denielle and Vilar, Mark Joshua},
title = {Deepfake Blackmailing on the Rise: The Burgeoning Posterity of Revenge Pornography in the Philippines},
year = {2024},
isbn = {9798400716713},
publisher = {Association for Computing Machinery},
address = {New York, NY, USA},
url = {https://doi.org/10.1145/3654522.3654548},
doi = {10.1145/3654522.3654548},
booktitle = {Proceedings of the 2024 9th International Conference on Intelligent Information Technology},
pages = {295–301},
numpages = {7},
location = {Ho Chi Minh City, Vietnam},
series = {ICIIT '24}
}

@article{Passos_2024,
   title={A review of deep learning‐based approaches for deepfake content detection},
   volume={41},
   ISSN={1468-0394},
   url={http://dx.doi.org/10.1111/EXSY.13570},
   DOI={10.1111/exsy.13570},
   number={8},
   journal={Expert Systems},
   publisher={Wiley},
   author={Passos, Leandro A. and Jodas, Danilo and Costa, Kelton A. P. and Souza Júnior, Luis A. and Rodrigues, Douglas and Del Ser, Javier and Camacho, David and Papa, João Paulo},
   year={2024},
   month=feb }

@ARTICLE{wang2004ssim,
  author={Zhou Wang and Bovik, A.C. and Sheikh, H.R. and Simoncelli, E.P.},
  journal={IEEE Transactions on Image Processing}, 
  title={Image quality assessment: from error visibility to structural similarity}, 
  year={2004},
  volume={13},
  number={4},
  pages={600-612},
  doi={10.1109/TIP.2003.819861}}

@ARTICLE{zhang2011fsim,
  author={Zhang, Lin and Zhang, Lei and Mou, Xuanqin and Zhang, David},
  journal={IEEE Transactions on Image Processing}, 
  title={FSIM: A Feature Similarity Index for Image Quality Assessment}, 
  year={2011},
  volume={20},
  number={8},
  pages={2378-2386},
  doi={10.1109/TIP.2011.2109730}
}

@inproceedings{radford2021clip,
  author       = {Alec Radford and
                  Jong Wook Kim and
                  Chris Hallacy and
                  Aditya Ramesh and
                  Gabriel Goh and
                  Sandhini Agarwal and
                  Girish Sastry and
                  Amanda Askell and
                  Pamela Mishkin and
                  Jack Clark and
                  Gretchen Krueger and
                  Ilya Sutskever},
  editor       = {Marina Meila and
                  Tong Zhang},
  title        = {Learning Transferable Visual Models From Natural Language Supervision},
  booktitle    = {Proceedings of the 38th International Conference on Machine Learning,
                  {ICML} 2021, 18-24 July 2021, Virtual Event},
  series       = {Proceedings of Machine Learning Research},
  volume       = {139},
  pages        = {8748--8763},
  publisher    = {{PMLR}},
  year         = {2021},
  url          = {http://proceedings.mlr.press/v139/radford21a.html},
  bibsource    = {dblp computer science bibliography, https://dblp.org}
}

@inproceedings{kirillov2023sam,
  title={Segment anything},
  author={Kirillov, Alexander and Mintun, Eric and Ravi, Nikhila and Mao, Hanzi and Rolland, Chloe and Gustafson, Laura and Xiao, Tete and Whitehead, Spencer and Berg, Alexander C and Lo, Wan-Yen and others},
  booktitle={Proceedings of the IEEE/CVF International Conference on Computer Vision},
  pages={4015--4026},
  year={2023}
}

@article{Zhou2018UNetAN,
  title={UNet++: A Nested U-Net Architecture for Medical Image Segmentation},
  author={Zongwei Zhou and Md Mahfuzur Rahman Siddiquee and Nima Tajbakhsh and Jianming Liang},
  journal={Deep Learning in Medical Image Analysis and Multimodal Learning for Clinical Decision Support : 4th International Workshop, DLMIA 2018, and 8th International Workshop, ML-CDS 2018, held in conjunction with MICCAI 2018, Granada, Spain, S...},
  year={2018},
  volume={11045},
  pages={
          3-11
        },
  url={https://api.semanticscholar.org/CorpusID:50786304}
}

@inproceedings{szegedy2014intriguingpropertiesneuralnetworks,
  author       = {Christian Szegedy and
                  Wojciech Zaremba and
                  Ilya Sutskever and
                  Joan Bruna and
                  Dumitru Erhan and
                  Ian J. Goodfellow and
                  Rob Fergus},
  editor       = {Yoshua Bengio and
                  Yann LeCun},
  title        = {Intriguing properties of neural networks},
  booktitle    = {2nd International Conference on Learning Representations, {ICLR} 2014,
                  Banff, AB, Canada, April 14-16, 2014, Conference Track Proceedings},
  year         = {2014},
  url          = {http://arxiv.org/abs/1312.6199},
  bibsource    = {dblp computer science bibliography, https://dblp.org}
}

@inbook{Biggio_2013,
   title={Evasion Attacks against Machine Learning at Test Time},
   ISBN={9783642387098},
   ISSN={1611-3349},
   url={http://dx.doi.org/10.1007/978-3-642-40994-3_25},
   DOI={10.1007/978-3-642-40994-3_25},
   booktitle={Advanced Information Systems Engineering},
   publisher={Springer Berlin Heidelberg},
   author={Biggio, Battista and Corona, Igino and Maiorca, Davide and Nelson, Blaine and Šrndić, Nedim and Laskov, Pavel and Giacinto, Giorgio and Roli, Fabio},
   year={2013},
   pages={387–402} }

@inproceedings{yang2014clothing,
  title={Clothing Co-Parsing by Joint Image Segmentation and Labeling},
  author={Yang, Wei and Luo, Ping and Lin, Liang},
  booktitle={Proceedings of the IEEE Conference on Computer Vision and Pattern Recognition (CVPR)},
  year={2014},
  organization={IEEE},
  note={Dataset available at \url{https://www.kaggle.com/datasets/balraj98/clothing-coparsing-dataset}}
}

@inproceedings{sohl2015deep,
  title={Deep unsupervised learning using nonequilibrium thermodynamics},
  author={Sohl-Dickstein, Jascha and Weiss, Eric and Maheswaranathan, Niru and Ganguli, Surya},
  booktitle={International conference on machine learning},
  pages={2256--2265},
  year={2015},
  organization={PMLR}
}

@article{ho2020denoising,
  title={Denoising diffusion probabilistic models},
  author={Ho, Jonathan and Jain, Ajay and Abbeel, Pieter},
  journal={Advances in neural information processing systems},
  volume={33},
  pages={6840--6851},
  year={2020}
}

@inproceedings{rombach2022high,
  title={High-resolution image synthesis with latent diffusion models},
  author={Rombach, Robin and Blattmann, Andreas and Lorenz, Dominik and Esser, Patrick and Ommer, Bj{\"o}rn},
  booktitle={Proceedings of the IEEE/CVF conference on computer vision and pattern recognition},
  pages={10684--10695},
  year={2022}
}

@inproceedings{brooks2023instructpix2pix,
  title={Instructpix2pix: Learning to follow image editing instructions},
  author={Brooks, Tim and Holynski, Aleksander and Efros, Alexei A},
  booktitle={Proceedings of the IEEE/CVF Conference on Computer Vision and Pattern Recognition},
  pages={18392--18402},
  year={2023}
}

@inproceedings{
couairon2023diffedit,
title={DiffEdit: Diffusion-based semantic image editing with mask guidance},
author={Guillaume Couairon and Jakob Verbeek and Holger Schwenk and Matthieu Cord},
booktitle={The Eleventh International Conference on Learning Representations },
year={2023},
url={https://openreview.net/forum?id=3lge0p5o-M-}
}

@inproceedings{
hertz2023prompttoprompt,
title={Prompt-to-Prompt Image Editing with Cross-Attention Control},
author={Amir Hertz and Ron Mokady and Jay Tenenbaum and Kfir Aberman and Yael Pritch and Daniel Cohen-or},
booktitle={The Eleventh International Conference on Learning Representations },
year={2023},
url={https://openreview.net/forum?id=_CDixzkzeyb}
}

@inproceedings{
meng2022sdedit,
title={{SDE}dit: Guided Image Synthesis and Editing with Stochastic Differential Equations},
author={Chenlin Meng and Yutong He and Yang Song and Jiaming Song and Jiajun Wu and Jun-Yan Zhu and Stefano Ermon},
booktitle={International Conference on Learning Representations},
year={2022},
url={https://openreview.net/forum?id=aBsCjcPu_tE}
}

@article{saharia2022photorealistic,
  title={Photorealistic text-to-image diffusion models with deep language understanding},
  author={Saharia, Chitwan and Chan, William and Saxena, Saurabh and Li, Lala and Whang, Jay and Denton, Emily L and Ghasemipour, Kamyar and Gontijo Lopes, Raphael and Karagol Ayan, Burcu and Salimans, Tim and others},
  journal={Advances in neural information processing systems},
  volume={35},
  pages={36479--36494},
  year={2022}
}

@inproceedings{ruiz2023dreambooth,
  title={Dreambooth: Fine tuning text-to-image diffusion models for subject-driven generation},
  author={Ruiz, Nataniel and Li, Yuanzhen and Jampani, Varun and Pritch, Yael and Rubinstein, Michael and Aberman, Kfir},
  booktitle={Proceedings of the IEEE/CVF conference on computer vision and pattern recognition},
  pages={22500--22510},
  year={2023}
}

@article{chefer2023attend,
  title={Attend-and-excite: Attention-based semantic guidance for text-to-image diffusion models},
  author={Chefer, Hila and Alaluf, Yuval and Vinker, Yael and Wolf, Lior and Cohen-Or, Daniel},
  journal={ACM Transactions on Graphics (TOG)},
  volume={42},
  number={4},
  pages={1--10},
  year={2023},
  publisher={ACM New York, NY, USA}
}

@inproceedings{zhang2023adding,
  title={Adding conditional control to text-to-image diffusion models},
  author={Zhang, Lvmin and Rao, Anyi and Agrawala, Maneesh},
  booktitle={Proceedings of the IEEE/CVF International Conference on Computer Vision},
  pages={3836--3847},
  year={2023}
}

@inproceedings{li2023gligen,
  title={Gligen: Open-set grounded text-to-image generation},
  author={Li, Yuheng and Liu, Haotian and Wu, Qingyang and Mu, Fangzhou and Yang, Jianwei and Gao, Jianfeng and Li, Chunyuan and Lee, Yong Jae},
  booktitle={Proceedings of the IEEE/CVF Conference on Computer Vision and Pattern Recognition},
  pages={22511--22521},
  year={2023}
}

@inproceedings{mou2024t2i,
  title={T2i-adapter: Learning adapters to dig out more controllable ability for text-to-image diffusion models},
  author={Mou, Chong and Wang, Xintao and Xie, Liangbin and Wu, Yanze and Zhang, Jian and Qi, Zhongang and Shan, Ying},
  booktitle={Proceedings of the AAAI Conference on Artificial Intelligence},
  volume={38},
  number={5},
  pages={4296--4304},
  year={2024}
}

@inproceedings{bansal2023universal,
  title={Universal guidance for diffusion models},
  author={Bansal, Arpit and Chu, Hong-Min and Schwarzschild, Avi and Sengupta, Soumyadip and Goldblum, Micah and Geiping, Jonas and Goldstein, Tom},
  booktitle={Proceedings of the IEEE/CVF Conference on Computer Vision and Pattern Recognition},
  pages={843--852},
  year={2023}
}

@article{segura2023jpeg,
  title={JPEG compressed images can bypass protections against ai editing},
  author={Sandoval-Segura, Pedro and Geiping, Jonas and Goldstein, Tom},
  journal={arXiv preprint arXiv:2304.02234},
  year={2023}
}

@inproceedings{wang2023stylediffusion,
  author       = {Zhizhong Wang and
                  Lei Zhao and
                  Wei Xing},
  title        = {StyleDiffusion: Controllable Disentangled Style Transfer via Diffusion
                  Models},
  booktitle    = {{IEEE/CVF} International Conference on Computer Vision, {ICCV} 2023,
                  Paris, France, October 1-6, 2023},
  pages        = {7643--7655},
  publisher    = {{IEEE}},
  year         = {2023},
  url          = {https://doi.org/10.1109/ICCV51070.2023.00706},
  doi          = {10.1109/ICCV51070.2023.00706},
  bibsource    = {dblp computer science bibliography, https://dblp.org}
}

@inproceedings{andreas2022repaint,
  author       = {Andreas Lugmayr and
                  Martin Danelljan and
                  Andr{\'{e}}s Romero and
                  Fisher Yu and
                  Radu Timofte and
                  Luc Van Gool},
  title        = {RePaint: Inpainting using Denoising Diffusion Probabilistic Models},
  booktitle    = {{IEEE/CVF} Conference on Computer Vision and Pattern Recognition,
                  {CVPR} 2022, New Orleans, LA, USA, June 18-24, 2022},
  pages        = {11451--11461},
  publisher    = {{IEEE}},
  year         = {2022},
  url          = {https://doi.org/10.1109/CVPR52688.2022.01117},
  doi          = {10.1109/CVPR52688.2022.01117},
  bibsource    = {dblp computer science bibliography, https://dblp.org}
}

@inproceedings{yang2023paint,
  author       = {Binxin Yang and
                  Shuyang Gu and
                  Bo Zhang and
                  Ting Zhang and
                  Xuejin Chen and
                  Xiaoyan Sun and
                  Dong Chen and
                  Fang Wen},
  title        = {Paint by Example: Exemplar-based Image Editing with Diffusion Models},
  booktitle    = {{IEEE/CVF} Conference on Computer Vision and Pattern Recognition,
                  {CVPR} 2023, Vancouver, BC, Canada, June 17-24, 2023},
  pages        = {18381--18391},
  publisher    = {{IEEE}},
  year         = {2023},
  url          = {https://doi.org/10.1109/CVPR52729.2023.01763},
  doi          = {10.1109/CVPR52729.2023.01763},
  bibsource    = {dblp computer science bibliography, https://dblp.org}
}

@inproceedings{zhang2023towards,
  author       = {Guanhua Zhang and
                  Jiabao Ji and
                  Yang Zhang and
                  Mo Yu and
                  Tommi S. Jaakkola and
                  Shiyu Chang},
  editor       = {Andreas Krause and
                  Emma Brunskill and
                  Kyunghyun Cho and
                  Barbara Engelhardt and
                  Sivan Sabato and
                  Jonathan Scarlett},
  title        = {Towards Coherent Image Inpainting Using Denoising Diffusion Implicit
                  Models},
  booktitle    = {International Conference on Machine Learning, {ICML} 2023, 23-29 July
                  2023, Honolulu, Hawaii, {USA}},
  series       = {Proceedings of Machine Learning Research},
  volume       = {202},
  pages        = {41164--41193},
  publisher    = {{PMLR}},
  year         = {2023},
  url          = {https://proceedings.mlr.press/v202/zhang23q.html},
  bibsource    = {dblp computer science bibliography, https://dblp.org}
}

@inproceedings{chong2024dragon,
  author       = {Chong Mou and
                  Xintao Wang and
                  Jiechong Song and
                  Ying Shan and
                  Jian Zhang},
  title        = {DragonDiffusion: Enabling Drag-style Manipulation on Diffusion Models},
  booktitle    = {The Twelfth International Conference on Learning Representations,
                  {ICLR} 2024, Vienna, Austria, May 7-11, 2024},
  publisher    = {OpenReview.net},
  year         = {2024},
  url          = {https://openreview.net/forum?id=OEL4FJMg1b},
  bibsource    = {dblp computer science bibliography, https://dblp.org}
}

@inproceedings{hertz2023delta,
  author       = {Amir Hertz and
                  Kfir Aberman and
                  Daniel Cohen{-}Or},
  title        = {Delta Denoising Score},
  booktitle    = {{IEEE/CVF} International Conference on Computer Vision, {ICCV} 2023,
                  Paris, France, October 1-6, 2023},
  pages        = {2328--2337},
  publisher    = {{IEEE}},
  year         = {2023},
  url          = {https://doi.org/10.1109/ICCV51070.2023.00221},
  doi          = {10.1109/ICCV51070.2023.00221},
  bibsource    = {dblp computer science bibliography, https://dblp.org}
}

@inproceedings{lin2024text,
  author       = {Yuanze Lin and
                  Yi{-}Wen Chen and
                  Yi{-}Hsuan Tsai and
                  Lu Jiang and
                  Ming{-}Hsuan Yang},
  title        = {Text-Driven Image Editing via Learnable Regions},
  booktitle    = {{IEEE/CVF} Conference on Computer Vision and Pattern Recognition,
                  {CVPR} 2024, Seattle, WA, USA, June 16-22, 2024},
  pages        = {7059--7068},
  publisher    = {{IEEE}},
  year         = {2024},
  url          = {https://doi.org/10.1109/CVPR52733.2024.00674},
  doi          = {10.1109/CVPR52733.2024.00674},
  bibsource    = {dblp computer science bibliography, https://dblp.org}
}

@inproceedings{guarav2023pixtopix,
  author       = {Gaurav Parmar and
                  Krishna Kumar Singh and
                  Richard Zhang and
                  Yijun Li and
                  Jingwan Lu and
                  Jun{-}Yan Zhu},
  editor       = {Erik Brunvand and
                  Alla Sheffer and
                  Michael Wimmer},
  title        = {Zero-shot Image-to-Image Translation},
  booktitle    = {{ACM} {SIGGRAPH} 2023 Conference Proceedings, {SIGGRAPH} 2023, Los
                  Angeles, CA, USA, August 6-10, 2023},
  pages        = {11:1--11:11},
  publisher    = {{ACM}},
  year         = {2023},
  url          = {https://doi.org/10.1145/3588432.3591513},
  doi          = {10.1145/3588432.3591513},
  bibsource    = {dblp computer science bibliography, https://dblp.org}
}

@inproceedings{guillaume2023diffedit,
  author       = {Guillaume Couairon and
                  Jakob Verbeek and
                  Holger Schwenk and
                  Matthieu Cord},
  title        = {DiffEdit: Diffusion-based semantic image editing with mask guidance},
  booktitle    = {The Eleventh International Conference on Learning Representations,
                  {ICLR} 2023, Kigali, Rwanda, May 1-5, 2023},
  publisher    = {OpenReview.net},
  year         = {2023},
  url          = {https://openreview.net/forum?id=3lge0p5o-M-},
  bibsource    = {dblp computer science bibliography, https://dblp.org}
}

@inproceedings{gwanghyun2022difclip,
  author       = {Gwanghyun Kim and
                  Taesung Kwon and
                  Jong Chul Ye},
  title        = {DiffusionCLIP: Text-Guided Diffusion Models for Robust Image Manipulation},
  booktitle    = {{IEEE/CVF} Conference on Computer Vision and Pattern Recognition,
                  {CVPR} 2022, New Orleans, LA, USA, June 18-24, 2022},
  pages        = {2416--2425},
  publisher    = {{IEEE}},
  year         = {2022},
  url          = {https://doi.org/10.1109/CVPR52688.2022.00246},
  doi          = {10.1109/CVPR52688.2022.00246},
  bibsource    = {dblp computer science bibliography, https://dblp.org}
}

@inproceedings{mokady2023null,
  author       = {Ron Mokady and
                  Amir Hertz and
                  Kfir Aberman and
                  Yael Pritch and
                  Daniel Cohen{-}Or},
  title        = {Null-text Inversion for Editing Real Images using Guided Diffusion
                  Models},
  booktitle    = {{IEEE/CVF} Conference on Computer Vision and Pattern Recognition,
                  {CVPR} 2023, Vancouver, BC, Canada, June 17-24, 2023},
  pages        = {6038--6047},
  publisher    = {{IEEE}},
  year         = {2023},
  url          = {https://doi.org/10.1109/CVPR52729.2023.00585},
  doi          = {10.1109/CVPR52729.2023.00585},
  bibsource    = {dblp computer science bibliography, https://dblp.org}
}

@inproceedings{ruiz2020disrupting,
  title={Disrupting deepfakes: Adversarial attacks against conditional image translation networks and facial manipulation systems},
  author={Ruiz, Nataniel and Bargal, Sarah Adel and Sclaroff, Stan},
  booktitle={Computer Vision--ECCV 2020 Workshops: Glasgow, UK, August 23--28, 2020, Proceedings, Part IV 16},
  pages={236--251},
  year={2020},
  organization={Springer}
}

@inproceedings{
madry2018towards,
title={Towards Deep Learning Models Resistant to Adversarial Attacks},
author={Aleksander Madry and Aleksandar Makelov and Ludwig Schmidt and Dimitris Tsipras and Adrian Vladu},
booktitle={International Conference on Learning Representations},
year={2018},
url={https://openreview.net/forum?id=rJzIBfZAb},
}

@inproceedings{li2023ntire_dn50, title={NTIRE 2023 Challenge on Image Denoising: Methods and Results}, author={Li, Yawei and Zhang, Yulun and Van Gool, Luc and Timofte, Radu and others}, booktitle={Proceedings of the IEEE/CVF Conference on Computer Vision and Pattern Recognition Workshops}, year={2023} }

@online{Prolific,
  author = {Prolific},
  title = {Prolific: Online participant recruitment for surveys and research},
  year = 2024,
  howpublished = {\url{https://prolific.com/}},
  note = {Accessed: 2024-11-01}
}

@online{chatgpt,
  author = {OpenAI},
  title = {ChatGPT},
  year = 2024,
  howpublished = {\url{https://chatgpt.com/}},
  note = {Accessed: 2024-10-02}
}

@article{daiki2023negative,
  title={Negative-prompt inversion: Fast image inversion for editing with text-guided diffusion models},
  author={Miyake, Daiki and Iohara, Akihiro and Saito, Yu and Tanaka, Toshiyuki},
  journal={arXiv preprint arXiv:2305.16807},
  year={2023}
}

@article{ravi2023preditor,
  title={Preditor: Text guided image editing with diffusion prior},
  author={Ravi, Hareesh and Kelkar, Sachin and Harikumar, Midhun and Kale, Ajinkya},
  journal={arXiv preprint arXiv:2302.07979},
  year={2023}
}

@inproceedings{kingma2015adam,
  author       = {Diederik P. Kingma and
                  Jimmy Ba},
  editor       = {Yoshua Bengio and
                  Yann LeCun},
  title        = {Adam: {A} Method for Stochastic Optimization},
  booktitle    = {3rd International Conference on Learning Representations, {ICLR} 2015,
                  San Diego, CA, USA, May 7-9, 2015, Conference Track Proceedings},
  year         = {2015},
  url          = {http://arxiv.org/abs/1412.6980},
  bibsource    = {dblp computer science bibliography, https://dblp.org}
}

@inproceedings{poursaeed2018generative,
  title={Generative adversarial perturbations},
  author={Poursaeed, Omid and Katsman, Isay and Gao, Bicheng and Belongie, Serge},
  booktitle={Proceedings of the IEEE conference on computer vision and pattern recognition},
  pages={4422--4431},
  year={2018}
}

@article{cao2023impress,
  title={Impress: Evaluating the resilience of imperceptible perturbations against unauthorized data usage in diffusion-based generative ai},
  author={Cao, Bochuan and Li, Changjiang and Wang, Ting and Jia, Jinyuan and Li, Bo and Chen, Jinghui},
  journal={Advances in Neural Information Processing Systems},
  volume={36},
  pages={10657--10677},
  year={2023}
}

@inproceedings{
choi2025diffusionguard,
title={DiffusionGuard: A Robust Defense Against Malicious Diffusion-based Image Editing},
author={June Suk Choi and Kyungmin Lee and Jongheon Jeong and Saining Xie and Jinwoo Shin and Kimin Lee},
booktitle={The Thirteenth International Conference on Learning Representations},
year={2025},
url={https://openreview.net/forum?id=9OfKxKoYNw}
}

@article{mist,
  title={Mist: Towards Improved Adversarial Examples for Diffusion Models},
  author={Liang, Chumeng and Wu, Xiaoyu},
  journal={arXiv preprint arXiv:2305.12683},
  year={2023}
}

@inproceedings{advdm,
  title={Adversarial example does good: Preventing painting imitation from diffusion models via adversarial examples},
  author={Liang, Chumeng and Wu, Xiaoyu and Hua, Yang and Zhang, Jiaru and Xue, Yiming and Song, Tao and Xue, Zhengui and Ma, Ruhui and Guan, Haibing},
  booktitle={International Conference on Machine Learning},
  pages={20763--20786},
  year={2023},
  organization={PMLR}
}

@inproceedings{
sds,
title={Toward effective protection against diffusion-based mimicry through score distillation},
author={Haotian Xue and Chumeng Liang and Xiaoyu Wu and Yongxin Chen},
booktitle={The Twelfth International Conference on Learning Representations},
year={2024},
url={https://openreview.net/forum?id=NzxCMe88HX}
}

@article{zhang2023magicbrush,
  title={Magicbrush: A manually annotated dataset for instruction-guided image editing},
  author={Zhang, Kai and Mo, Lingbo and Chen, Wenhu and Sun, Huan and Su, Yu},
  journal={Advances in Neural Information Processing Systems},
  volume={36},
  pages={31428--31449},
  year={2023}
}

@inproceedings{liupseudo,
  title={Pseudo Numerical Methods for Diffusion Models on Manifolds},
  author={Liu, Luping and Ren, Yi and Lin, Zhijie and Zhao, Zhou},
  booktitle={International Conference on Learning Representations}
}

@article{karras2022elucidating,
  title={Elucidating the design space of diffusion-based generative models},
  author={Karras, Tero and Aittala, Miika and Aila, Timo and Laine, Samuli},
  journal={Advances in neural information processing systems},
  volume={35},
  pages={26565--26577},
  year={2022}
}

@article{guo2024greybox,
  title={A Grey-box Attack against Latent Diffusion Model-based Image Editing by Posterior Collapse},
  author={Guo, Zhongliang and Lei, Chun Tong and Fang, Lei and Zhao, Shuai and Qian, Yifei and Lin, Jingyu and Wang, Zeyu and Chen, Cunjian and Arandjelovi{\'c}, Ognjen and Lau, Chun Pong},
  journal={arXiv preprint arXiv:2408.10901},
  year={2024}
}

@article{chen2023editshield,
  title={EditShield: Protecting Unauthorized Image Editing by Instruction-guided Diffusion Models},
  author={Chen, Ruoxi and Jin, Haibo and Liu, Yixin and Chen, Jinyin and Wang, Haohan and Sun, Lichao},
  journal={arXiv preprint arXiv:2311.12066},
  year={2023}
}

@article{shih2025pixel,
  title={Pixel Is Not a Barrier: An Effective Evasion Attack for Pixel-Domain Diffusion Models},
  author={Shih, Chun-Yen and Peng, Li-Xuan and Liao, Jia-Wei and Chu, Ernie and Chou, Cheng-Fu and Chen, Jun-Cheng},
  journal={arXiv preprint arXiv:2408.11810},
  year={2025}
}
\bibliographystyle{iclr2026_conference}

\appendix

\newpage
\appendix

\section{Appendix}

\setcounter{tocdepth}{4}
\localtableofcontents






\newpage
\subsection{Model Algorithm and Implementation Details}
\label{a:imp-details}

\begin{figure}[ht]
\centering
\includegraphics[width=0.9\textwidth]{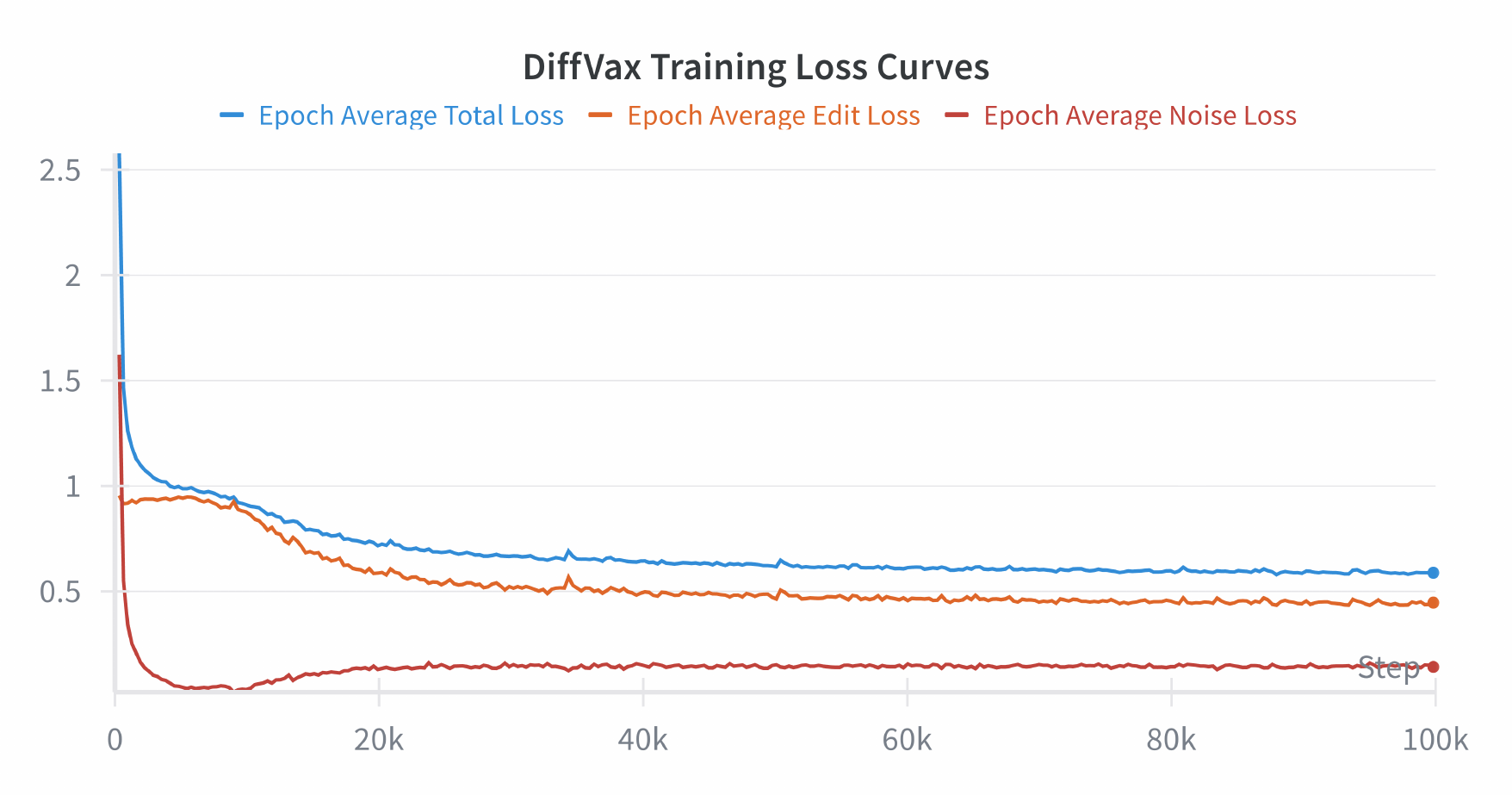}
\caption{\textit{\textbf{Training Loss Curves.}} We report the epoch-averaged Total Loss (blue), Edit Loss (orange), and Noise Loss (red). The curves demonstrate that the noise loss drops quickly to a low magnitude, ensuring the perturbation remains imperceptible, while the edit loss reduces continuously as the immunizer optimizes for edit disruption.}
\label{fig:loss-curve}
\end{figure}

\paragraph{Implementation Details}
We employ a UNet++ architecture~\citep{Zhou2018UNetAN} for the immunizer model. We selected this over a standard U-Net because its nested skip pathways facilitate denser feature aggregation at different semantic levels. Empirically, we found that this dense connectivity provides significantly better training stability for the unstable optimization task of predicting adversarial noise, allowing the model to generate more precise, imperceptible high-frequency perturbations. We train our immunizer model for 350 epochs using a batch size of 5 on an NVIDIA A100 GPU. We use the Adam optimizer~\citep{kingma2015adam} with an initial learning rate of 0.00001 and set the loss weight parameter $\alpha = 4$. Training takes approximately 22 hours and leverages 16-bit precision to reduce memory consumption and speed up computation. As illustrated in Figure~\ref{fig:loss-curve}, the training process exhibits high stability; the noise loss converges rapidly to ensure imperceptibility, while the edit loss decreases steadily as the model progressively learns to disrupt the editing process. For the editing tools, we use a pre-trained Stable Diffusion v1.5 inpainting model~\citep{rombach2022high} for inpainting-based editing, and InstructPix2Pix~\citep{brooks2023instructpix2pix} for instruction-based editing tasks.

\paragraph{Training Algorithm}
Algorithm~\ref{alg:cap} describes the end-to-end training procedure for our immunizer model. For each data sample, the model generates an immunized image by injecting noise into the masked region. This image is then edited using a black-box editing model. The training objective minimizes both the deviation from the original image in the masked region and the effectiveness of the edit in the unmasked region.

\begin{center}
\begin{minipage}{0.7\linewidth}
\begin{algorithm}[H]
\caption{End-to-end Training Framework}
\label{alg:cap}
\hspace*{\algorithmicindent} \textbf{Input:} Immunizer model $\unet{\cdot}$, Editing model $\diffusion{\cdot}$, Dataset $\dataset$, Dataset size $N$, Loss weight $\alpha$
\begin{algorithmic}
\For{$n = 1$ to $N$}
\State $(\image^n, \mask^n, \prompt^n) \gets \text{sample}(\dataset, n)$
\State $\immunizationnoise^n \gets \unet{\image^n}$
\State $\immunizedimage^n \gets (\image^n + \immunizationnoise^n \odot \mask^n).\text{clamp}(0,1)$
\State $\editedimage^n \gets \diffusion{\immunizedimage^n, \sim\mask^n, \prompt^n}$
\State $\noiseloss \gets \text{normalize}(\|(\immunizedimage^n - \image^n) \odot \mask^n\|_1)$
\State $\editloss \gets \text{normalize}(\|\editedimage^n \odot (\sim \mask^n)\|_1)$
\State $\totalloss \gets \alpha \cdot \noiseloss + \editloss$
\State $\unetparameters \gets \text{update}(\nabla_{\unetparameters} \totalloss)$
\EndFor
\end{algorithmic}
\end{algorithm}
\end{minipage}
\end{center}

\paragraph{Dataset Setup}
\label{a:dataset-setup}
Our dataset consists of 1,000 images, each associated with two prompts, resulting in a total of 2,000 prompts. We split the dataset into 80\% for the training set (seen) and 20\% for the validation set (unseen). The prompt set was constructed using ChatGPT~\citep{chatgpt}, specifically by generating prompts designed for background editing. A total of 1,000 prompts were collected and subsequently split into 80\% for the training set (seen) and 20\% for the validation set (unseen). Finally, we sampled two random prompts for each image in the dataset, ensuring the prompts corresponded to whether the image was categorized as seen or unseen.

Our dataset is comparable in size to the current datasets used in related works, and is therefore aligned with the current standard of evidence in the field, while more data would always be better. To place our dataset size in the context of prior work, the closest research for training a generative adversarial noise generator is the paper "Generative Adversarial Perturbations"~\citep{poursaeed2018generative}. For their experiments on semantic segmentation, they used the focused Cityscapes dataset, which contains 2,975 training and 500 validation images. Given that this foundational work was established on a dataset of a few thousand images from a specific domain (urban scenes), we believe our dataset of 875 human images is in a comparable range for a proof-of-concept study. Nevertheless, we believe that extending our method to larger and more diverse datasets is a crucial next step, and we will highlight this as an important avenue for future work.

\newpage
\subsection{Additional Qualitative Results and Comparisons}
\label{sec:qualitative_results}

\subsubsection{Additional Results with Inpainting-Based Editing Models}
\begin{figure*}[ht]
\centering
\includegraphics[width=\textwidth]{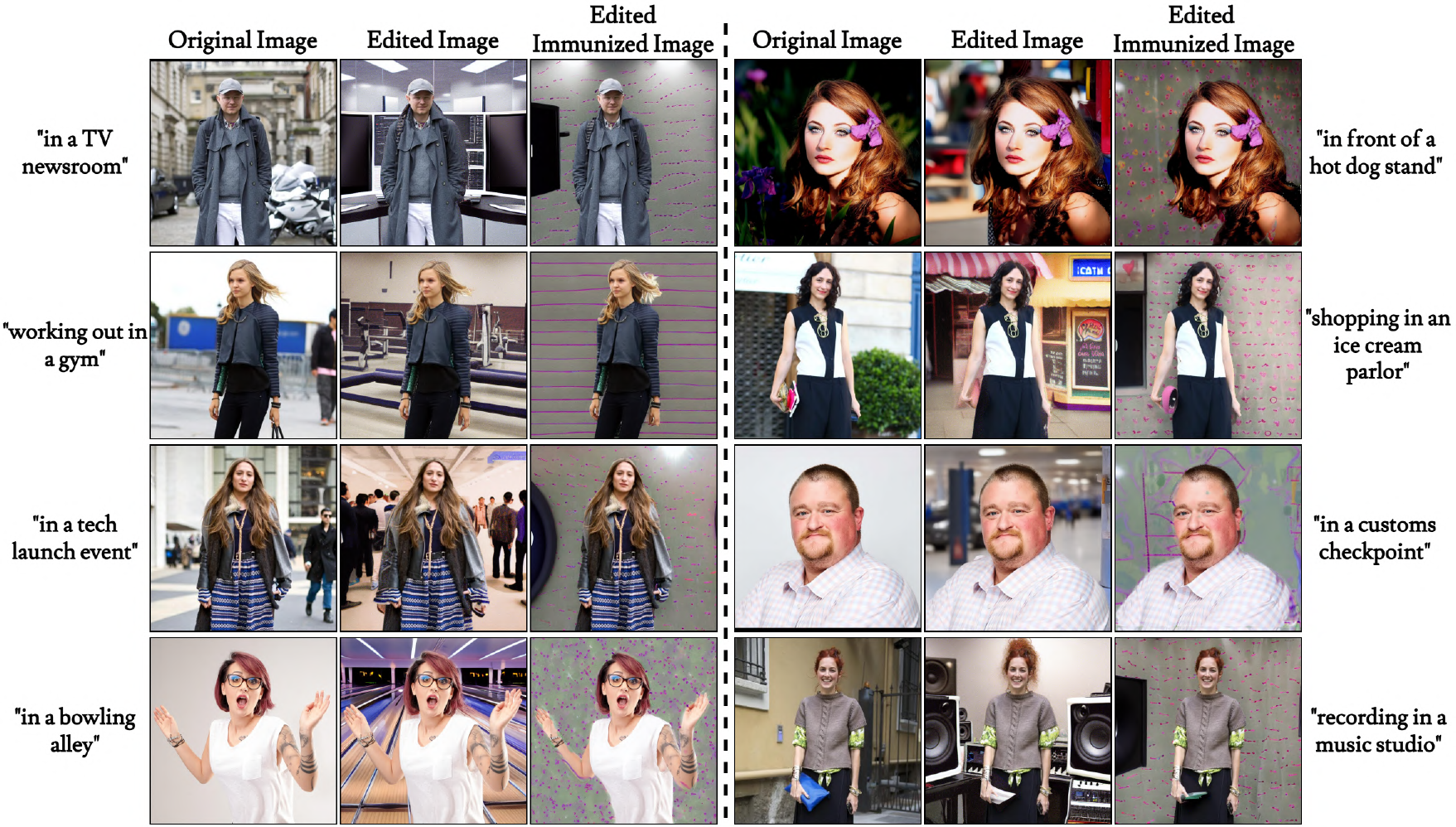}
\caption{\textit{\textbf{Additional qualitative results with \methodname.}} Each row displays a different prompt and input image, illustrating \methodname's ability to consistently disrupt harmful edits. Despite varying and challenging prompts, the edited outputs from the protected images show clear signs of disruption, emphasizing the robustness of our method.}
\label{fig:additional_qualitative}
\end{figure*}

Figure~\ref{fig:additional_qualitative} presents supplementary qualitative results obtained using inpainting-based editing models. The examples cover a wide range of scenarios and prompts, demonstrating the effectiveness of our immunization method on previously unseen content. Notably, the model performs well even on close-up images, maintaining robustness against malicious edits in both broad and fine-grained contexts.

\newpage
\subsubsection{Additional Comparisons with Inpainting-Based Editing Models}

\begin{figure*}[ht]
\centering
\includegraphics[width=\textwidth]{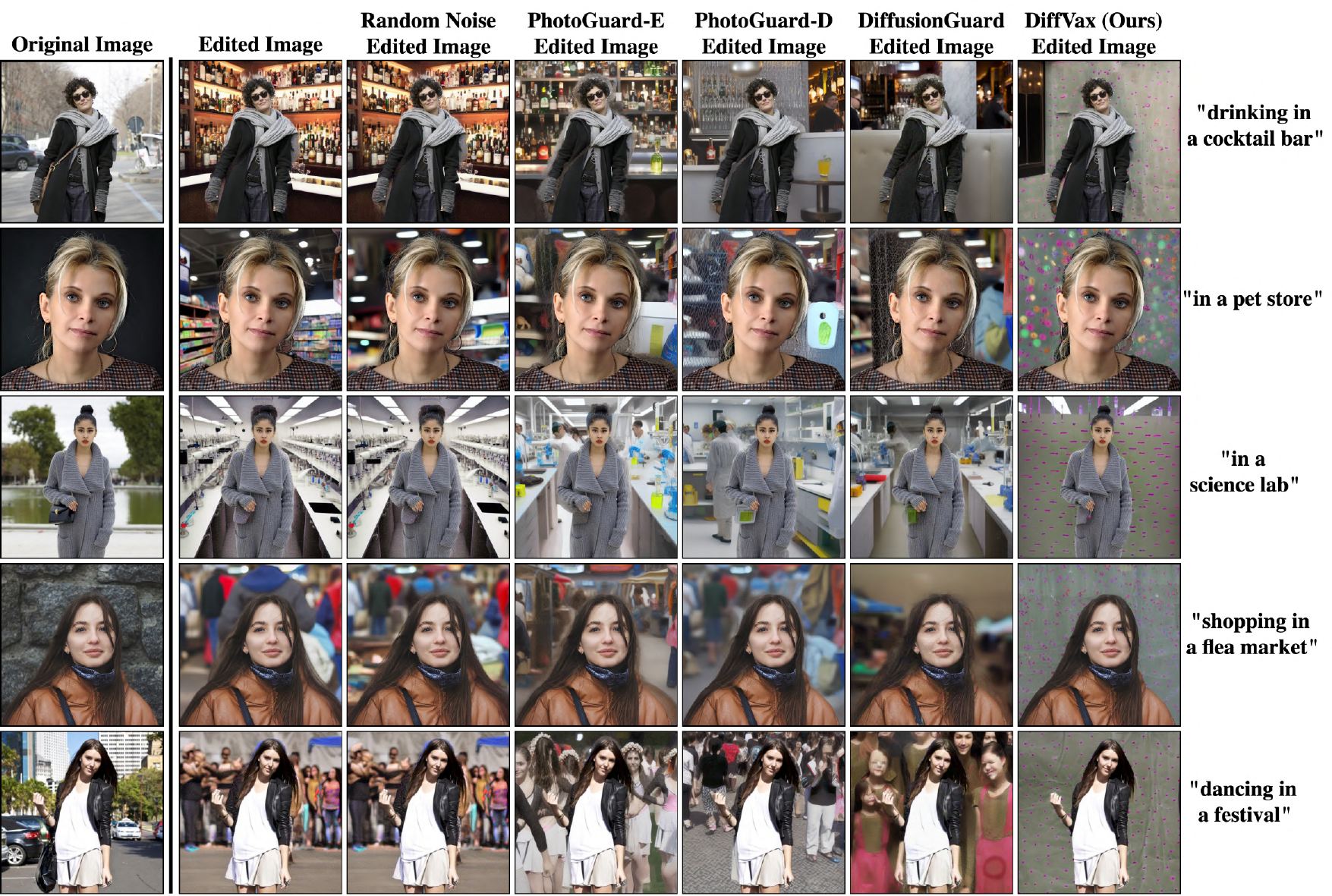}
\caption{\textit{\textbf{Additional qualitative comparison between baselines and \methodname.}} Each row represents a unique prompt-image pair, while the columns show outputs for different immunization methods. \methodname\ consistently produces better results, effectively disrupting edits while preserving image quality.}
\label{fig:additional_qualitative_comparison}
\end{figure*}

Figure~\ref{fig:additional_qualitative_comparison} shows extended qualitative comparisons between \methodname\ and various baseline immunization methods, including Random Noise, PhotoGuard-E, PhotoGuard-D, and DiffusionGuard. These results are produced using inpainting-based editing models. The comparison highlights how \methodname\ consistently achieves better performance in visually disrupting malicious edits while preserving the semantic integrity of the original image.

We note that other defense methods such as AdvDM~\citep{advdm}, SDS~\citep{sds}, and Mist~\citep{mist} have also been proposed in the literature. However, these techniques are tailored for specific editing pipelines like SDEdit~\citep{meng2022sdedit} and are not directly applicable in our inpainting-based setup, thus making direct comparison beyond our experimental scope.

\newpage
\subsubsection{Additional Results with Instruction-Based Editing Model}

\begin{figure*}[!ht]
\centering
\includegraphics[width=0.8\textwidth]{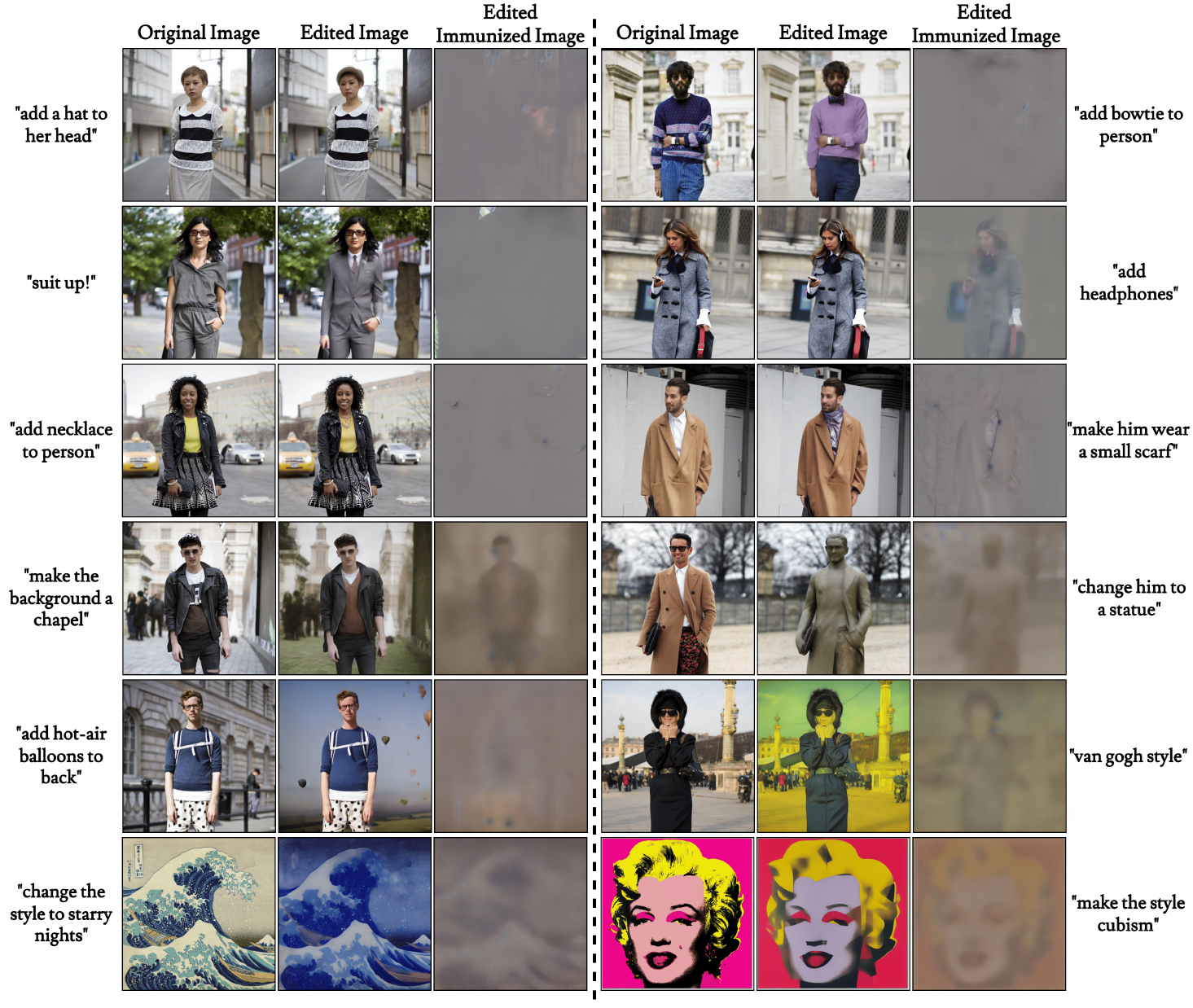}
\caption{\textit{\textbf{Qualitative results using the InstructPix2Pix~\citep{brooks2023instructpix2pix} editing model with \methodname.}} Each triplet shows an original image, its edited counterpart, and the result after immunization. \methodname\ successfully prevents a diverse set of edits, including background replacement, style transfer, object insertion, and attribute modification, further demonstrating its generalizability across editing types.}
\label{fig:qualitative_ip2p}
\end{figure*}

To further evaluate the generalizability of \methodname, we apply it to edits generated using InstructPix2Pix~\citep{brooks2023instructpix2pix}, a widely adopted text-guided diffusion-based editing tool. This setting differs significantly from inpainting models, as edits are applied based on high-level natural language instructions.
As shown in Figure~\ref{fig:qualitative_ip2p}, \methodname\ consistently disrupts a broad range of editing intents across various image types. The examples illustrate the model's robustness across:

\begin{itemize}
    \item \textbf{Human attribute edits} (e.g., \textit{"add a hat to her head"}, \textit{"add bowtie to person"}, \textit{"make him wear a small scarf"}): \methodname\ suppresses the addition of these features, effectively neutralizing changes to facial and clothing attributes.
    
    \item \textbf{Background edits} (e.g., \textit{"make the background a chapel"}, \textit{"change him to a statue"}): Despite significant changes to the scene, the edits fail to render properly on immunized images, showcasing \methodname's ability to neutralize edits in large non-focal areas.
    
    \item \textbf{Style transfer edits} (e.g., \textit{"change the style to starry nights"}, \textit{"make the style cubism"}, \textit{"van gogh style"}): \methodname\ prevents global transformations from taking effect, demonstrating its efficacy in blocking even abstract stylistic alterations.
    
    \item \textbf{Non-ROI edits} (e.g., \textit{"add hot-air balloons to back"}, \textit{"add necklace to person"}, \textit{"add headphones"}): These involve subtle object insertions in the background or around the subject. Even though the modification targets are not directly in the immunized region, \methodname\ still effectively disrupts the edit.
\end{itemize}

These results validate the model-agnostic and instruction-resilient nature of \methodname, confirming its applicability to both local and global edit intents.

\newpage
\subsubsection{Additional Evaluation with MagicBrush and Other Editing Models}

The landscape of generative editing models is vast and rapidly evolving. Our choice of evaluation models was guided by established benchmarks in the image immunization literature to ensure a fair and direct comparison with prior state-of-the-art methods. To further strengthen our claims of generalizability, we conducted an additional experiment comparing our approach against PhotoGuard on the modern, training-free editing model MagicBrush. As shown in Table~\ref{tab:magicbrush-comparison}, our learned perturbations remain effective at disrupting edits, demonstrating that the protection generalizes beyond the standard inpainting and instruction-based models used in prior benchmarks. Our preliminary results show that \methodname\ achieves superior edit disruption (lower SSIM, PSNR, FSIM, and CLIP-T) with comparable imperceptibility (SSIM Noise).

\begin{table*}[ht]
\centering
\caption{\small{\textit{\textbf{MagicBrush Comparison}}}}
\resizebox{0.8\textwidth}{!}{%
\begin{tabular}{c c c c c c}
\toprule
\textbf{MagicBrush} & SSIM $\downarrow$ & PSNR $\downarrow$ & FSIM $\downarrow$ & CLIP-T $\downarrow$ & SSIM (Noise) $\uparrow$ \\
\midrule
PhotoGuard & 0.682 & 18.81 & 0.546 & 25.64 & \textbf{0.967} \\
DiffVax    & \textbf{0.635} & \textbf{18.41} & \textbf{0.529} & \textbf{22.18} & 0.965 \\
\bottomrule
\end{tabular}
}
\label{tab:magicbrush-comparison}
\end{table*}

Table~\ref{tab:editing-models} contextualizes our evaluation scope by comparing the editing tools used across recent immunization works. The scope of our evaluation is aligned with current best practices. We acknowledge that methods like Prompt-to-Prompt~\cite{hertz2023prompttoprompt} and Null-text inversion~\cite{mokady2023null} represent a different editing paradigm not directly compatible with the current experimental setup, and adapting our framework to protect against them is a promising direction for future work.

\begin{table*}[ht]
\centering
\caption{\small{\textit{\textbf{Editing Models Used}}}}
\resizebox{0.7\textwidth}{!}{%
\begin{tabular}{l l}
\toprule
\textbf{Method} & \textbf{Editing Models Used} \\
\midrule
\textbf{DiffVax (Ours)}             & SD Inpainting, IP2P, MagicBrush \\
\textbf{DiffusionGuard~\cite{choi2025diffusionguard} (ICLR 2025)}  & SD Inpainting, IP2P \\
\textbf{PhotoGuard~\cite{photoguard} (ICML 2023)}     & SD Inpainting, SDEdit \\
\textbf{SDS~\cite{sds} (ICLR 2024)}             & SDEdit, SD Inpainting, Textual inversion \\
\textbf{Mist~\cite{mist} (ICML 2023)}             & Textual inversion, Dreambooth \\
\textbf{AdvDM~\cite{advdm} (ICML 2023)}             & Textual inversion, SDEdit \\
\bottomrule
\end{tabular}
}
\label{tab:editing-models}
\end{table*}

\newpage
\subsection{Additional Robustness Evaluations and Studies}
\label{a:robustness}

\subsubsection{Robustness to Different Sampling Steps and Sampler Settings}

To evaluate the robustness of \methodname\ against attackers who may vary their inference-time settings, we conduct experiments with different sampling steps and diffusion samplers. The results, presented in Table~\ref{tab:sampling-step-comparison} and Table~\ref{tab:sampler-comparison}, demonstrate that \methodname\ consistently and effectively disrupts malicious edits across a range of configurations.

Table~\ref{tab:sampling-step-comparison} shows that \methodname\ maintains superior performance across various sampling step counts (10, 20, 30, and 40). In nearly all scenarios, it achieves the best (lowest) scores for PSNR, FSIM, and CLIP-T, indicating its protection is not compromised when an attacker uses fewer or more steps for generation. Similarly, Table~\ref{tab:sampler-comparison} illustrates that \methodname\ outperforms baselines when different samplers (PNDMScheduler~\cite{liupseudo}, EulerDiscreteScheduler, LMSDiscreteScheduler~\cite{karras2022elucidating}) are used. This confirms that our learned immunization is not overfitted to a specific generation algorithm and remains effective in diverse, real-world attack scenarios.

\begin{table*}[ht]
\centering
\caption{\small{\textit{\textbf{Sampling Step Comparison}}}}
\begin{tabular}{cccccc}
\toprule
\textbf{Sampling Step} & \textbf{Model} & SSIM $\downarrow$ & PSNR $\downarrow$ & FSIM $\downarrow$ & CLIP-T $\downarrow$ \\
\midrule
    & PG-D        & 0.637 & 16.79 & 0.391 & 26.54 \\
 10 & DiffusionGuard   & 0.651 & 16.65 & 0.409 & 23.84 \\
    & DiffVax     & \textbf{0.627} & \textbf{16.37} & \textbf{0.366} & \textbf{22.96} \\
\midrule
  & PG-D        & \textbf{0.564} & 15.56 & 0.379 & 28.89 \\
20 & DiffusionGuard   & 0.591 & 15.28 & 0.393 & 26.04 \\
    & DiffVax     & \textbf{0.564} & \textbf{14.96} & \textbf{0.360} & \textbf{24.42} \\
\midrule
    & PG-D        & \textbf{0.523} & 14.92 & 0.379 & 29.27 \\ 30 & DiffusionGuard   & 0.556 & 14.71 & 0.386 & 27.10 \\
    & DiffVax     & 0.526 & \textbf{14.32} & \textbf{0.362} & \textbf{24.17} \\
\midrule
 & PG-D        & \textbf{0.507} & 14.42 & 0.377 & 29.68 \\
  40 & DiffusionGuard   & 0.539 & 14.16 & 0.386 & 27.84 \\
    & DiffVax     & 0.506 & \textbf{13.78} & \textbf{0.356} & \textbf{24.06} \\
\bottomrule
\end{tabular}
\label{tab:sampling-step-comparison}
\end{table*}

\begin{table*}[ht]
\centering
\caption{\small{\textit{\textbf{Sampler Comparison}}}}
\begin{tabular}{c c c c c c}
\toprule
\textbf{Sampler} & \textbf{Model} & SSIM $\downarrow$ & PSNR $\downarrow$ & FSIM $\downarrow$ & CLIP-T $\downarrow$ \\
\midrule
          & PG-D      & 0.480 & 14.31 & 0.404 & 26.88 \\
    PNDMScheduler        & DiffusionGuard & 0.501 & 14.52 & 0.404 & 26.97 \\
                       & DiffVax   & \textbf{0.440} & \textbf{13.41} & \textbf{0.372} & \textbf{21.67} \\
\midrule
 & PG-D      & 0.504 & 14.93 & 0.399 & 28.08 \\
    EulerDiscreteScheduler        & DiffusionGuard & 0.530 & 14.93 & 0.406 & 27.28 \\
                       & DiffVax   & \textbf{0.466} & \textbf{13.91} & \textbf{0.361} & \textbf{22.00} \\
\midrule
    & PG-D      & 0.487 & 14.36 & 0.403 & 27.82 \\
    LMSDiscreteScheduler        & DiffusionGuard & 0.509 & 14.47 & 0.405 & 27.23 \\
                       & DiffVax   & \textbf{0.449} & \textbf{13.43} & \textbf{0.367} & \textbf{21.70} \\
\bottomrule
\end{tabular}
\label{tab:sampler-comparison}
\end{table*}

\newpage

\subsubsection{Immunization Noise Comparison Under Perturbation Budget}
\label{a:pert-budget}
We have run experiments comparing DiffVax's average learned perturbation against baselines with fixed 16/255, 32/255, 64/255 budgets, and we performed evaluation based on the mean magnitude ($L_1$) of the immunization noise (perturbation). 

The results clearly show that DiffVax achieves superior edit disruption with a much smaller mean magnitude ($L_1$) perturbation than baselines given a larger budget, highlighting that its strength lies in the strategic placement of noise, not simply its magnitude. This supports our claim that DiffVax learns a more efficient and targeted noise distribution rather than applying uniform, high-energy noise. 

Unlike methods that enforce a rigid and uniform $L_p$ budget, DiffVax implicitly learns the perturbation's properties via the trade-off in our loss function, $\mathcal{L} = \alpha \cdot \mathcal{L}_{\text{noise}} + \mathcal{L}_{\text{edit}}$. This allows the model to strategically allocate its "budget," applying stronger noise only where most effective and least visible.

\begin{table*}[ht]
\centering
\caption{\small{\textit{\textbf{Comparison Across Immunization Strengths ($\epsilon$)}}}}
\resizebox{0.95\textwidth}{!}{%
\begin{tabular}{c c c c c c c c}
\toprule
\boldmath$\epsilon$ & \textbf{Method} & SSIM $\downarrow$ & PSNR $\downarrow$ & FSIM $\downarrow$ & CLIP-T $\downarrow$ & SSIM (Noise) $\uparrow$ & Mean Magnitude (L1) of Immunization Noise $\downarrow$ \\
\midrule
64/255 & PG-D       & \textbf{0.492} & 14.13 & 0.355 & 27.85 & 0.947 & 0.007 \\
       & DiffusionGuard  & 0.507 & 13.98 & 0.360 & 24.83 & 0.900 & 0.012 \\
\midrule
32/255 & PG-D       & 0.502 & 14.23 & 0.360 & 29.18 & 0.950 & 0.006 \\
       & DiffusionGuard  & 0.526 & 14.30 & 0.373 & 26.13 & 0.927 & 0.009 \\
\midrule
16/255 & PG-D       & 0.528 & 14.60 & 0.387 & 30.27 & 0.978 & 0.003 \\
       & DiffusionGuard  & 0.546 & 14.46 & 0.388 & 26.36 & 0.965 & 0.005 \\
\midrule
--     & DiffVax    & 0.496 & \textbf{13.85} & \textbf{0.352} & \textbf{22.96} & \textbf{0.989} & \textbf{0.001} \\
\bottomrule
\end{tabular}
}
\label{tab:epsilon-comparison}
\end{table*}

\newpage
\subsubsection{Robustness to Counterattacks}
\label{sec:robustness_counterattacks}

JPEG compression and denoising techniques are typically designed to remove high-frequency components from images. Since our immunizer model introduces primarily low-frequency perturbations—due to the design of our noise loss—it becomes inherently more robust against such counterattacks. 

Table~\ref{tab:additional_counterattack} reports results under various JPEG compression ratios and when using IMPRESS~\citep{cao2023impress}, a model specifically developed for adversarial purification and denoising. Across all configurations, \methodname\ consistently outperforms PhotoGuard-D and DiffusionGuard in terms of SSIM, SSIM (Noise), and CLIP-T metrics. These results suggest that \methodname\ maintains its protective efficacy even when subjected to aggressive counterattack scenarios.

\begin{figure*}[ht]
\centering
\includegraphics[width=0.8\columnwidth]{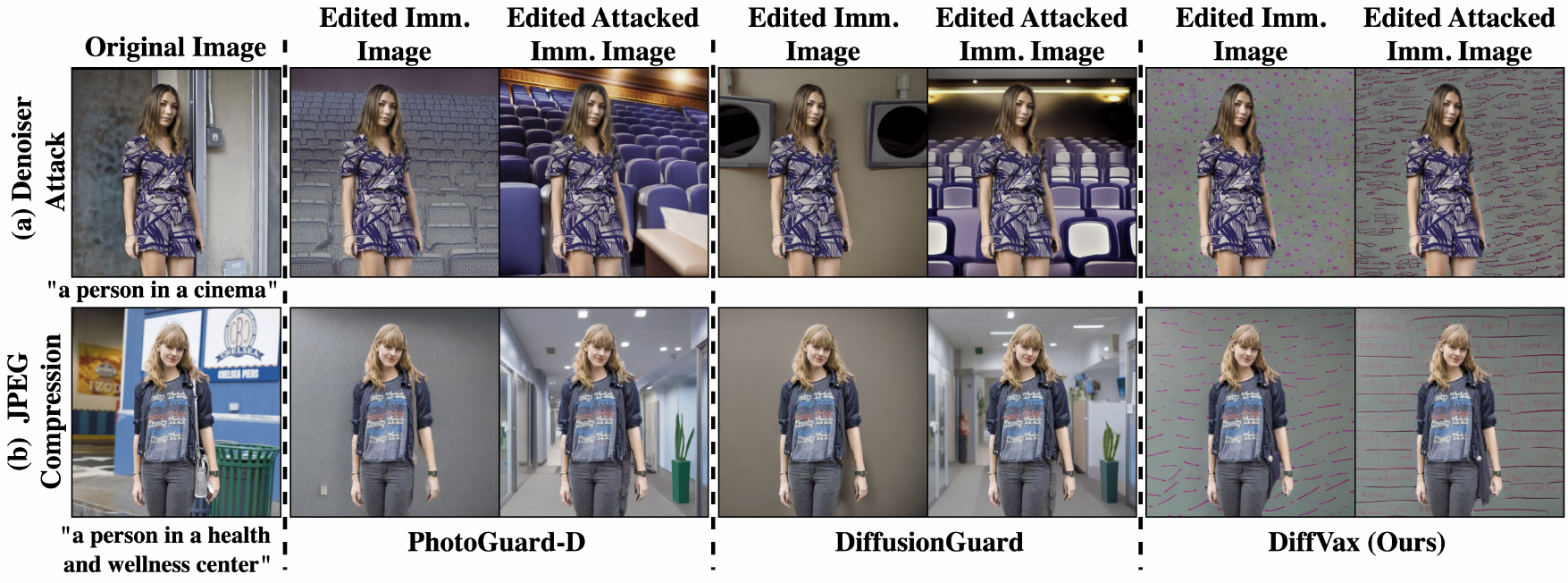} 
\caption{\textit{\textbf{Qualitative results of counter-attacks on immunization methods.}} The first row shows results when an off-the-shelf denoiser is applied to the immunized image, while the second row displays results under JPEG compression. Columns 2--3 correspond to PhotoGuard-D, while columns 4--5 show results for \methodname. PhotoGuard-D is visibly more susceptible to counterattacks, whereas \methodname\ maintains strong protection.}
\label{fig:imm-attack}
\end{figure*}

Figure~\ref{fig:imm-attack} presents qualitative results of two counterattack strategies: (a) applying a denoiser and (b) applying JPEG compression. The edited image, along with its attacked counterpart, is shown for both PhotoGuard-D and \methodname. While the visual changes for PhotoGuard-D are significant—indicating its vulnerability to counterattacks—\methodname\ retains its robustness, preventing successful malicious edits.

To further explore robustness, Figure~\ref{fig:imm-attack-2} presents additional qualitative comparisons under varying JPEG compression ratios (from 0.85 to 0.55) and under the IMPRESS purification attack. Even at high compression levels, \methodname\ continues to disrupt the edits, showcasing its superior generalization and resistance to counter-editing.

\begin{table*}[h]
\centering
\caption{\textbf{\textit{Additional counterattack experiments.}} The SSIM, SSIM (Noise), and CLIP-T metrics are reported for JPEG compression with ratios of 0.85, 0.65, and 0.55, as well as for the adversarial purification model IMPRESS. The metrics demonstrate that \methodname\ consistently outperforms PhotoGuard-D (PG) and DiffusionGuard (DG), even when counterattacks are applied to all methods.}
\resizebox{\textwidth}{!}{%
\begin{tabular}{c|ccc|ccc|ccc|ccc}
\toprule
  Metric & \shortstack{\methodname \\ (JPEG .85)} & \shortstack{DG \\ (JPEG .85)} & \shortstack{PG \\ (JPEG .85)} & \shortstack{\methodname \\ (JPEG .65)} & \shortstack{DG \\ (JPEG .65)} & \shortstack{PG \\ (JPEG .65)} & \shortstack{\methodname \\ (JPEG .55)} & \shortstack{DG \\ (JPEG .55)} & \shortstack{PG \\ (JPEG .55)} & \shortstack{\methodname \\ (IMPRESS)} & \shortstack{DG \\ (IMPRESS)} & \shortstack{PG \\ (IMPRESS)} \\ \midrule
SSIM $\downarrow$ & \textbf{0.517} & 0.646 & 0.640 & \textbf{0.530} & 0.696 & 0.692 & \textbf{0.534} & 0.706 & 0.693 & \textbf{0.489} & 0.605 & 0.578 \\
SSIM (Noise) $\uparrow$ & \textbf{0.968} & 0.955 & 0.961 & \textbf{0.951} & 0.946 & 0.950 & \textbf{0.944} & 0.940 & 0.944 & \textbf{0.644} & 0.636 & 0.640 \\ 
CLIP-T $\downarrow$ & \textbf{25.76} & 30.83 & 32.00 & \textbf{26.83} & 31.80 & 32.15 & \textbf{27.67} & 31.93 & 32.20 & \textbf{24.67} & 30.71 & 31.35 \\ \bottomrule
\end{tabular}%
}
\label{tab:additional_counterattack}
\end{table*}

\newpage

\begin{figure*}[ht]
\centering
\includegraphics[width=0.8\columnwidth]{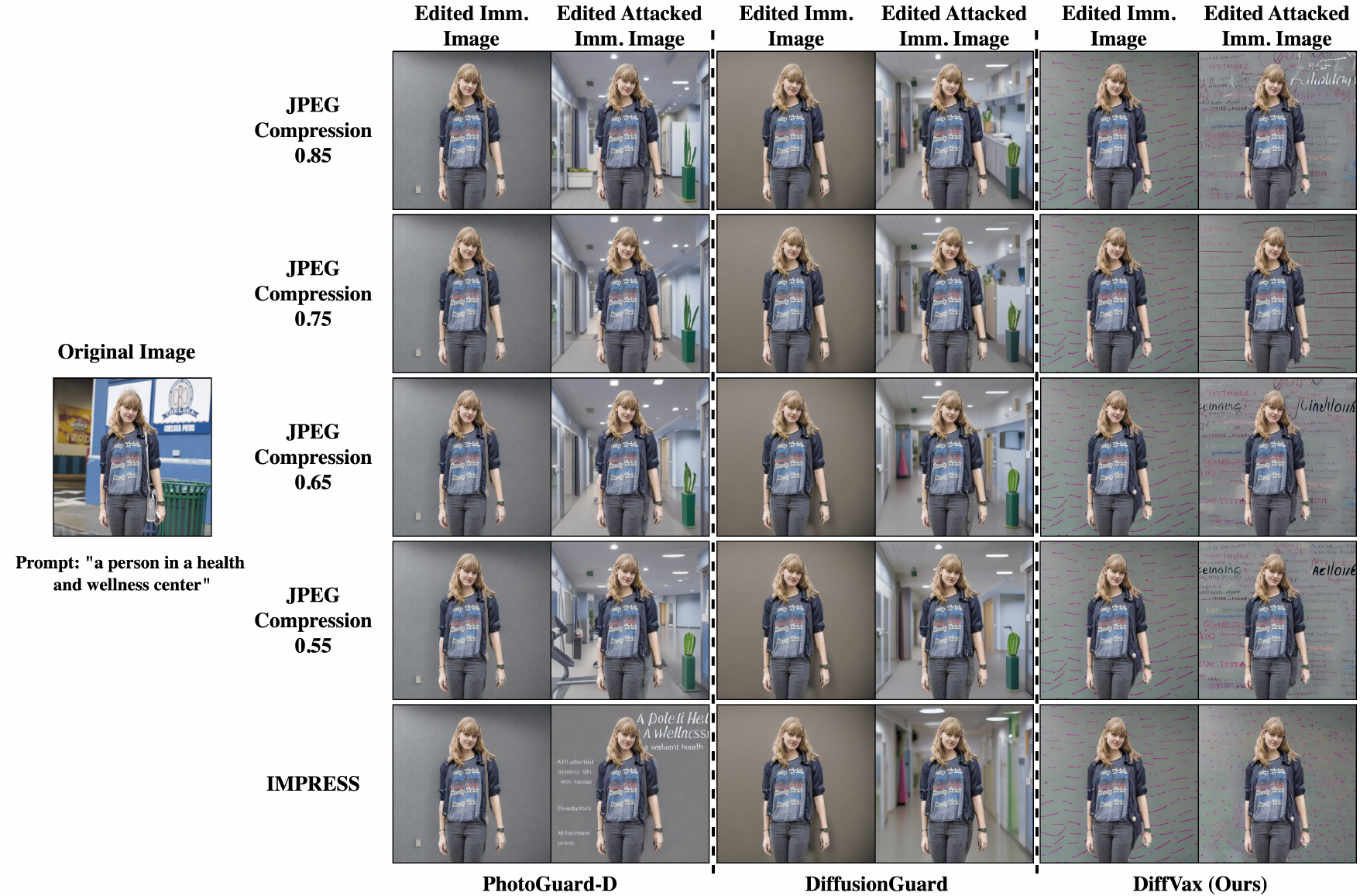} 
\caption{\textit{\textbf{Additional qualitative results of counter-attacks on immunization methods.}} Each row corresponds to a different JPEG compression ratio or the IMPRESS model. \methodname\ shows robust behavior across all levels, continuing to suppress harmful edits even under heavy degradation or purification.}
\label{fig:imm-attack-2}
\end{figure*}

\newpage
\subsubsection{Robustness to Non-Human Subjects}

\begin{figure*}[ht]
\centering
\includegraphics[width=0.6\textwidth]{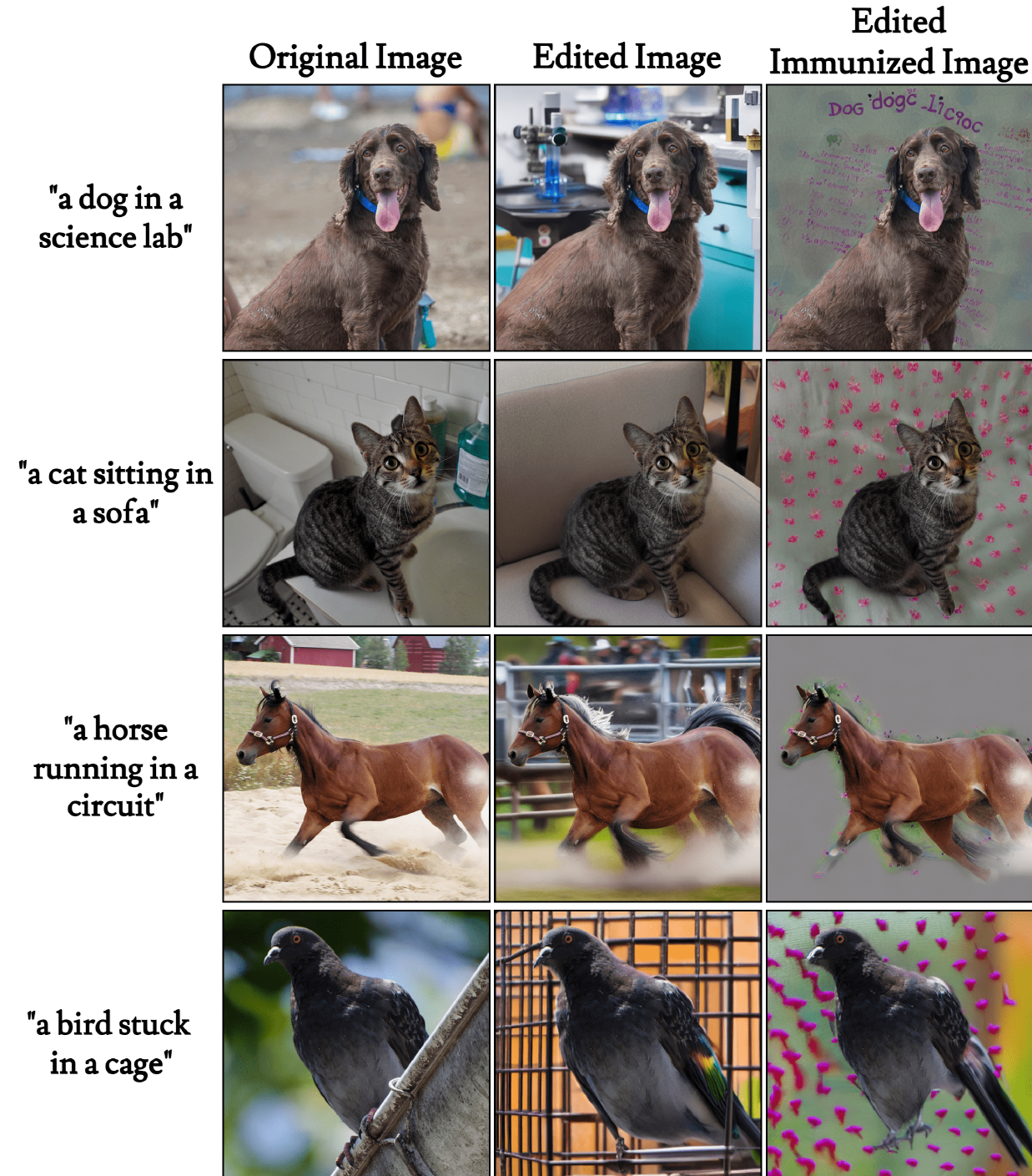}
\caption{\textit{\textbf{Qualitative results for non-human objects edited using \methodname.}} These examples show that \methodname\ extends effectively to domains beyond human subjects, maintaining its edit-resistance and imperceptibility.}
\label{fig:qualitative_other_objects}
\end{figure*}

To evaluate the generalizability of \methodname\ beyond human-centric content, we conduct experiments on non-human subjects, such as animals and other inanimate objects. As illustrated in Figure~\ref{fig:qualitative_other_objects}, \methodname\ effectively immunizes these non-person regions, preventing malicious edits while preserving the visual fidelity of the original image. These results further demonstrate the versatility and zero-shot capabilities of \methodname\ across diverse object domains.

\newpage
\subsubsection{User Study}
\label{sec:user_study}

To assess the human-perceived quality and effectiveness of each immunization method, we conducted a user study with 67 participants recruited via Prolific. Participants were asked to rank edited images based on how unrealistic or misaligned they appeared.

\begin{figure}[ht]
\centering
\includegraphics[width=0.4\columnwidth]{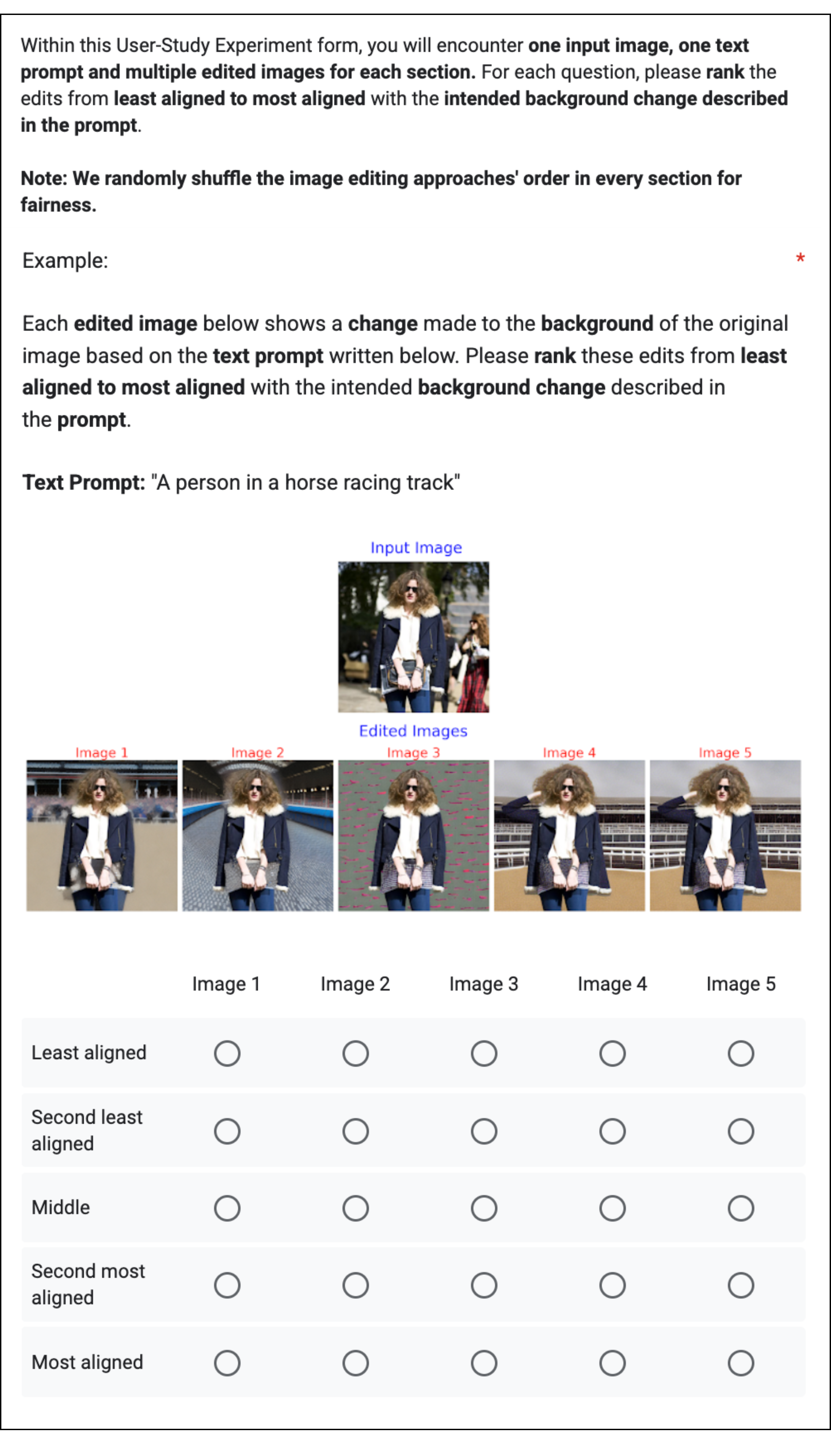}
\caption{\textit{\textbf{Instructions provided to user study participants.}} Users were asked to rank edited images from least to most aligned with the text prompt. Lower alignment suggests more successful immunization.}
\label{fig:user_study}
\end{figure}

Each participant was shown a set of five edited images derived from the same input image and text prompt (see Figure~\ref{fig:user_study}). These five outputs corresponded to different immunization strategies: Random Noise, PhotoGuard-E, PhotoGuard-D, \methodname, and an unprotected baseline. For each prompt-image pair, participants were instructed to rank the edits from \textbf{least aligned} to \textbf{most aligned} with the editing prompt. A lower ranking indicates better disruption of the intended edit (i.e., more effective immunization), as participants found the result less realistic or aligned with the prompt.

We randomly shuffled the order of methods in each trial to avoid position bias. In total, the study included 20 image-prompt pairs covering both seen and unseen examples, ensuring a fair and comprehensive evaluation.

\begin{table*}[ht]
\centering
\caption{\small{\textit{\textbf{User Study Rankings.}} Lower values indicate better perceived editing failure prevention, imperceptibility, and alignment with the original content.}}
\resizebox{0.5\textwidth}{!}{%
\begin{tabular}{l|c}
\toprule
\textbf{Immunization Method} & \textbf{Average Ranking} $\downarrow$ \\
\midrule
Random Noise & 3.74 \\
PhotoGuard-E & 3.33 \\
PhotoGuard-D & \underline{\textit{2.63}} \\
\cellcolor{green!25} \methodname\ (Ours) & \cellcolor{green!25} \textbf{1.64} \\
\bottomrule
\end{tabular}}
\label{tab:human_study}
\end{table*}

As shown in Table~\ref{tab:human_study}, \methodname\ significantly outperforms prior methods, receiving the best average ranking of \textbf{1.64}. This demonstrates the effectiveness of our method in fooling editing models in a way that is perceptually convincing to human observers. The next-best method, PhotoGuard-D, trails behind with a score of 2.63, while other methods rank even lower.

\newpage
\subsubsection{Video Evaluation}
\label{sec:video_eval}

To our knowledge, this is the first immunization-based video evaluation using a diffusion model for editing. We construct a video benchmark consisting of 4 human activity videos, each containing 64 frames and paired with 4 unique prompts. Since no prior method directly supports training-free video immunization using inpainting-based diffusion models, we adopt a naive per-frame editing pipeline to extend our approach to video. Despite not incorporating any explicit temporal modeling, our method yields strong results.

As reported in Table~\ref{tab:results-video}, \methodname\ outperforms all baselines across multiple metrics, including PSNR, SSIM (Noise), CLIP-T, and runtime. Notably, it achieves a dramatic reduction in runtime—processing the full dataset in just \textbf{0.739 seconds}—compared to PhotoGuard-D’s 64-hour runtime. These results emphasize the efficiency and practicality of our approach in real-time or large-scale settings.

Importantly, we make no architectural or training modifications for video data. The strong results achieved without temporal modeling suggest that our method generalizes well across sequential data, capturing consistent patterns in human identity, pose, and structure across frames. This robustness is further demonstrated in Fig.~\ref{fig:teaser} and Fig.~\ref{fig:qual} (c), where the model effectively adapts to changes in body motion and facial expressions.

Our work targets general-purpose editing protection and is evaluated on diverse, open-domain video data. The effectiveness of our approach under such settings demonstrates its promise as a scalable and general immunization strategy for future video editing systems.

\begin{table*}[ht]
\centering
\caption{\small{\textit{\textbf{Results on video editing.}} We report the average PSNR, SSIM, FSIM, SSIM (Noise), CLIP-T, and total runtime for Random Noise, PhotoGuard-D, DiffusionGuard, and \methodname\ on a video dataset consisting of 4 videos, each with 4 prompts and 64 frames. Best results per column are \textbf{bolded}.}}
\resizebox{0.95\textwidth}{!}{%
\begin{tabular}{lcccccc}
\toprule
\textbf{Method} & SSIM $\downarrow$ & PSNR $\downarrow$ & FSIM $\downarrow$ & SSIM (Noise) $\uparrow$ & CLIP-T $\downarrow$ & Runtime $\downarrow$ \\
\midrule
Random Noise      & 0.774 & 21.09 & 0.547 & 0.786 & 29.62 & N/A \\
PhotoGuard-D      & 0.738 & 17.31 & 0.448 & 0.965 & 26.52 & 64 hours \\
DiffusionGuard    & 0.750 & 17.43 & 0.478 & 0.922 & 25.41 & 10 hours \\
\methodname       & \textbf{0.681} & \textbf{16.78} & \textbf{0.374} & \textbf{0.974} & \textbf{22.51} & \textbf{0.739 seconds} \\
\bottomrule
\end{tabular}
}
\label{tab:results-video}
\end{table*}

\newpage
\subsection{Discussion on Generalization}
\label{sec:generalization}

While a universal immunizer that works zero-shot across all editing model architectures is a challenging open problem, \methodname\ demonstrates superior generalization compared to existing optimization-based methods across three distinct dimensions: generalization to unseen models, to unseen content, and to unseen masks. This section details these advantages.

\subsubsection{Generalization to Unseen Models}
\label{a:unseen-models}

\begin{figure*}[ht]
    \centering
    \includegraphics[width=\textwidth]{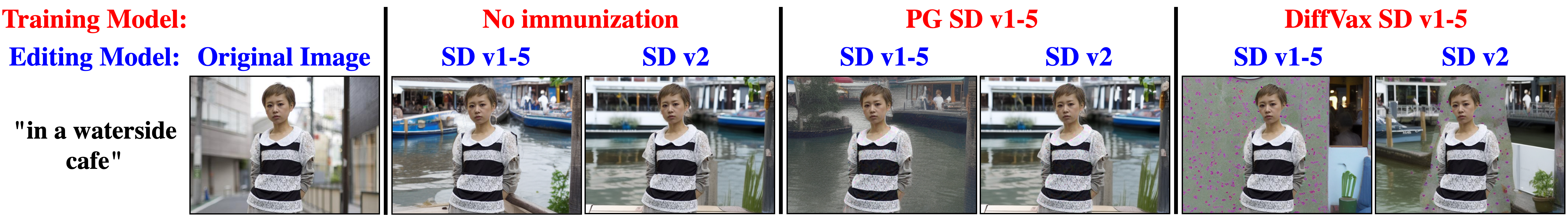}
\caption{\textit{\textbf{Transferability of perturbations across editing models.}}
    Red labels indicate the immunization training model, and blue labels denote the editing model. The results show how well each immunized image resists edits across different model configurations. When trained on Stable Diffusion (SD) v1.5, \methodname\ successfully prevents edits even when tested on SD v2. In contrast, PhotoGuard’s perturbations trained on SD v1.5 do not generalize to SD v2. These results illustrate the superior cross-model generalizability of \methodname.}
    \label{fig:generalization-comp}
\end{figure*}

Existing immunization methods, including optimization-based approaches like PhotoGuard, are model-specific. While developing a universally transferable immunizer is not the primary focus of this work, \methodname\ demonstrates significantly better generalization to unseen models than prior methods. We conducted an experiment where immunization noise was generated using a model trained on Stable Diffusion (SD) v1.5 and then tested on an unseen SD v2 model. As shown qualitatively in Figure~\ref{fig:generalization-comp}, \methodname\ successfully transfers its protective effect, whereas PhotoGuard's perturbations fail completely, leaving the image vulnerable.

Table~\ref{tab:transferability} provides quantitative results for this black-box transfer task, confirming that \methodname\ achieves the best performance across all metrics. This provides direct evidence that our learned immunization strategy is more robust and generalizable across model versions than optimization-based approaches.

\begin{table}[h]
\centering
\caption{Quantitative results for transferring immunization from SD v1.5 to an unseen SD v2.0 model. Lower values are better for all metrics, indicating more effective edit disruption. \methodname\ outperforms all baselines.}
\label{tab:transferability}
\begin{tabular}{lcccc}
\toprule
\textbf{SD 2.0} & \textbf{SSIM $\downarrow$} & \textbf{PSNR $\downarrow$} & \textbf{FSIM $\downarrow$} & \textbf{CLIP-T $\downarrow$} \\
\midrule
PG-D & 0.566 & 15.17 & 0.417 & 32.00 \\
DiffusionGuard & 0.609 & 15.26 & 0.454 & 31.73 \\
\textbf{\methodname} & \textbf{0.540} & \textbf{14.02} & \textbf{0.384} & \textbf{27.72} \\
\bottomrule
\end{tabular}
\end{table}

\subsubsection{Generalization to Unseen Content}
\label{a:unseen-content}

Optimization-based methods inherently handle unseen images by running a costly, per-image optimization process. A key scientific question this paper addresses is whether it is possible to learn a single feed-forward model that can directly generate effective perturbations without optimization. The success of our approach implies that the set of effective perturbations across all possible images possesses sufficient structure and regularity to be learnable. Our experiments demonstrate that \methodname\ successfully generalizes to \textbf{unseen images, unseen prompts, and even unseen videos} with a single forward pass, as demonstrated in Fig.~\ref{fig:teaser} and Fig.~\ref{fig:qual} (b) and (c) and Table~\ref{tab:results-video}. This establishes the learnability of the perturbation set for the first time and enables protection at a scale and speed previously unattainable.

\newpage
\subsubsection{Generalization to Unseen Masks during Test Time}
\label{a:unseen-masking}

Most existing state-of-the-art (SOTA) methods assume that the same mask is used during both the immunization (training) and editing (testing) phases. While this assumption aligns with standardized deepfake pipelines—where masks are often fixed to cover specific regions such as the head or full body—it limits the robustness of these methods to real-world scenarios involving unpredictable or mismatched editing masks.

\begin{figure}[ht]
\centering
\includegraphics[width=\textwidth]{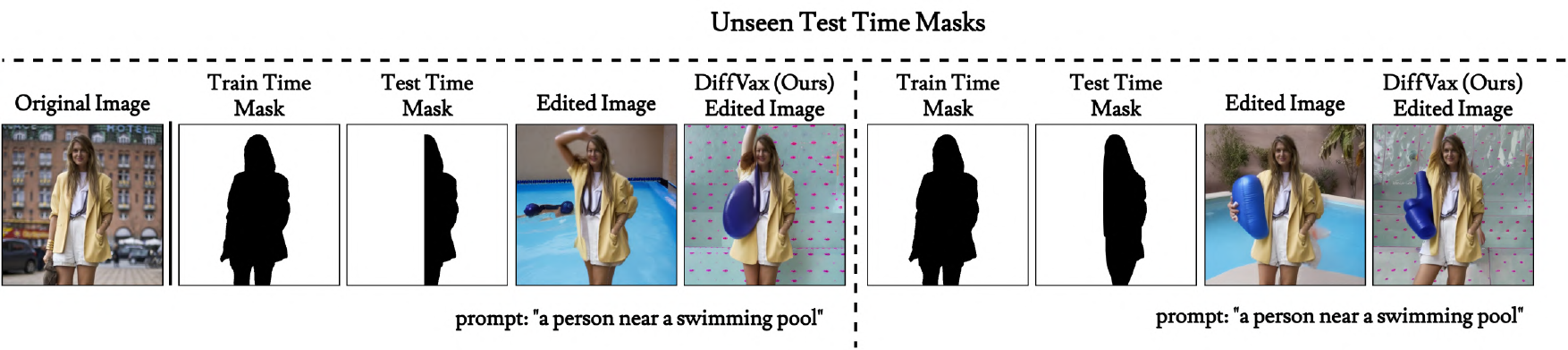}
\caption{\textit{\textbf{Additional comparison of partially covering masks.}} We evaluate robustness in scenarios where the test-time mask only covers a portion of the subject (e.g., a vertical half-split) compared to the full mask used during immunization. As shown, \methodname\ successfully maintains protection and disrupts the edit despite the partial mask coverage.}
\label{fig:half-mask}
\end{figure}

\begin{figure*}[ht]
    \centering
    \includegraphics[width=0.7\textwidth]{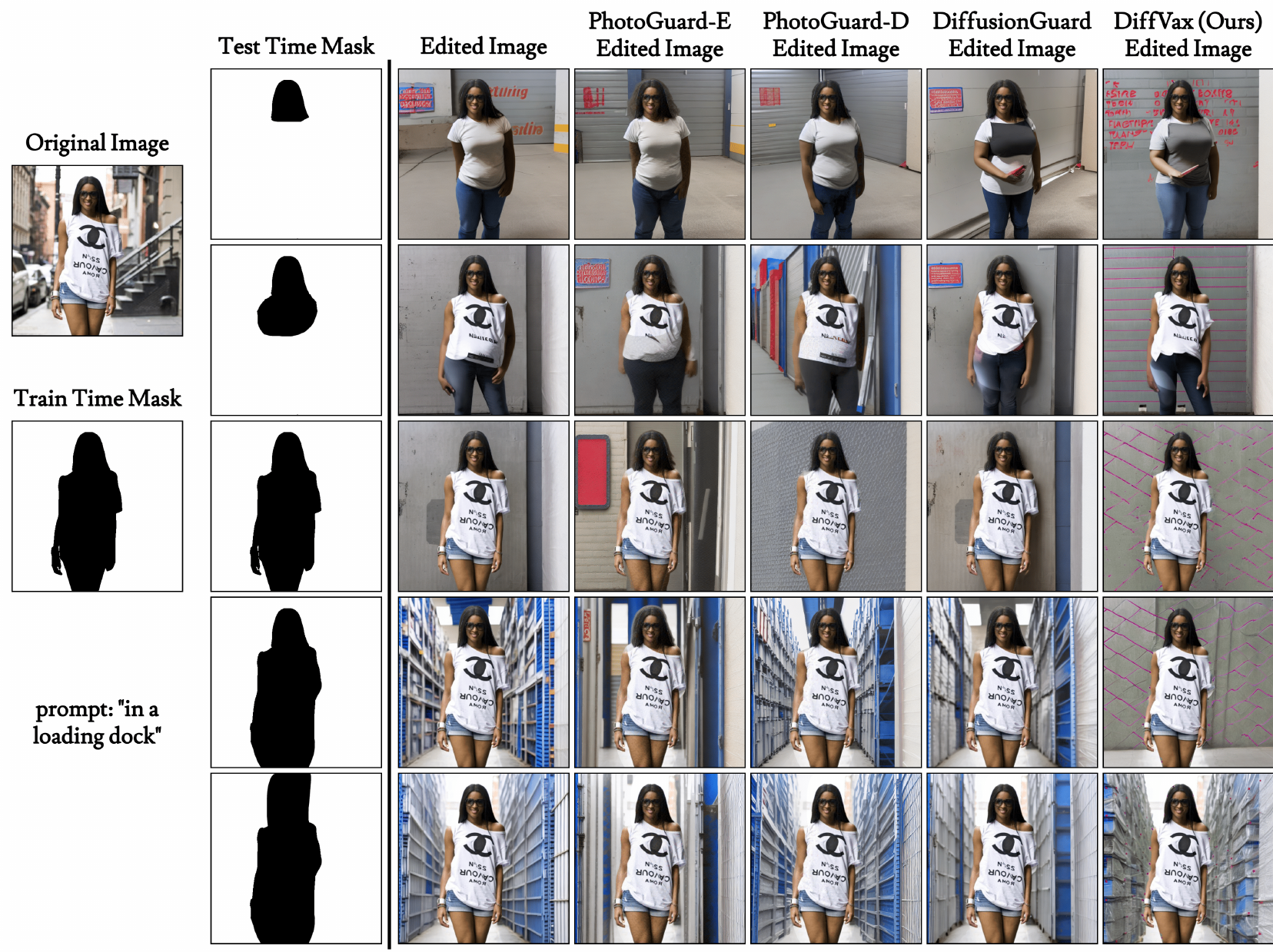}
    \caption{\textit{\textbf{Comparison of edited immunized images with different immunization and editing masks.}}
    PhotoGuard uses the same mask for both training and testing, making it highly sensitive to changes in the editing mask. \methodname, by contrast, is trained with a fixed immunization mask but remains robust even when the test-time editing mask significantly deviates. The results show consistent disruption of edits by \methodname\ despite large mask variability.}
    \label{fig:test-time-masks}
\end{figure*}

To evaluate this limitation, we conduct an experiment where the editing mask during test time differs from the mask used during immunization. As shown in Figure~\ref{fig:test-time-masks}, when the test-time mask diverges from the training mask, existing methods such as PhotoGuard (PG) and DiffusionGuard fail to maintain their edit-disrupting behavior. In contrast, \methodname\ remains effective, successfully disrupting the malicious edits even when significant changes are made to the mask size or region. This robustness can be attributed to our model’s design, which does not overfit to the spatial shape or scale of the mask used during training. Instead, it learns to encode more generalizable perturbations that degrade editing attempts across a range of editing contexts. These findings suggest that \methodname\ offers better real-world applicability where attackers may alter masks to evade immunization. We further demonstrate this in Figure~\ref{fig:half-mask}, which illustrates that \methodname\ maintains its protective capabilities even when the test-time mask only partially covers the subject (e.g., a vertical half-mask), preventing the edit despite the minimal mask overlap.

\newpage
\subsection{Imperceptibility Discussion}
\label{a:impercep}
\begin{figure*}[ht]
\centering
\includegraphics[width=0.5\textwidth]{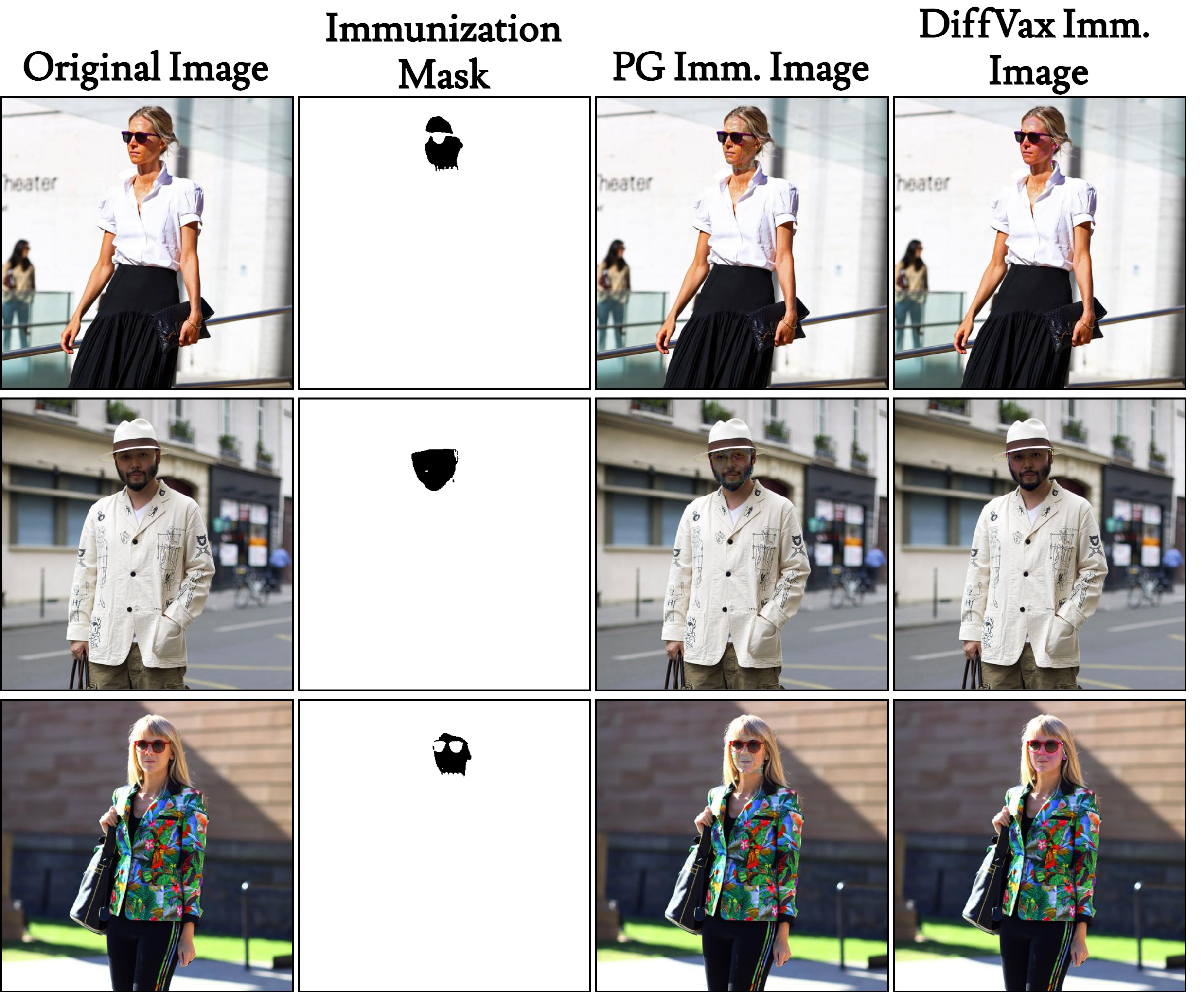}
\caption{\textit{\textbf{Comparison of immunization noise.}}  
Visual comparison of immunized images generated by PhotoGuard and \methodname\ using a face mask. PhotoGuard produces scattered and higher-frequency noise, while \methodname\ generates smoother, low-frequency perturbations.}
\label{fig:imm-2}
\end{figure*}

To evaluate the imperceptibility of the perturbations introduced by \methodname, we present qualitative comparisons against PhotoGuard in Figure~\ref{fig:imm-2}. Our method generates noise that is concentrated in the low-frequency components of the image, making it visually more subtle and less disruptive. In contrast, PhotoGuard introduces high-frequency noise that appears scattered across broader regions.

This low-frequency characteristic of \methodname\ offers two key advantages. First, it enhances the perceptual quality of the immunized images by producing smoother perturbations that minimally interfere with semantic content. Second, it contributes to robustness against counterattacks such as JPEG compression or denoising—these techniques are typically designed to suppress high-frequency information, which is assumed to correspond to noise. Since \methodname\ avoids relying on high-frequency artifacts, its perturbations are more likely to survive such transformations, preserving the protective effect.

We further examine the role of the loss norm in shaping the visual quality of the immunization. As shown in Figure~\ref{fig:imm-attack-2}, using $L_2$ or $L_\infty$ norms leads to less perceptible perturbations than the default $L_1$ formulation. However, this comes at the expense of reduced edit resistance, underscoring a critical trade-off between imperceptibility and robustness.

Future work will explore more principled approaches to navigating this trade-off, such as incorporating perceptual similarity metrics or frequency-domain regularization directly into the optimization objective.

\begin{figure*}[ht]
\centering
\includegraphics[width=0.9\textwidth]{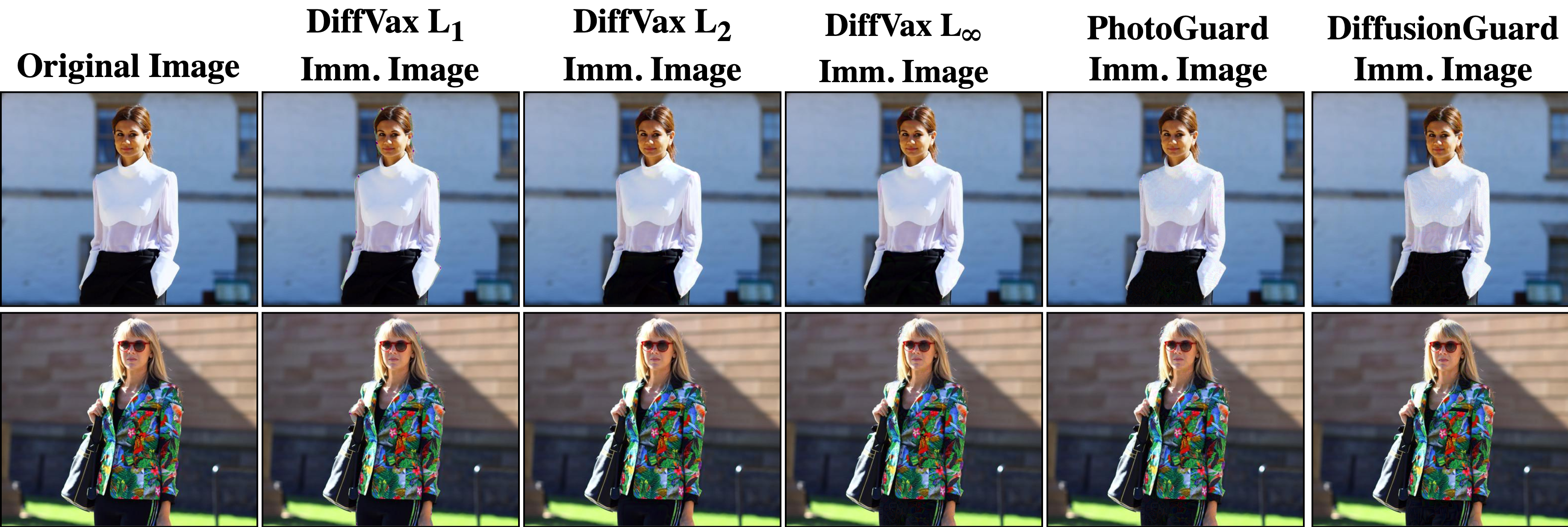}
\caption{\textit{\textbf{Additional comparison of immunization noise under different norms.}}  
This figure compares immunized images generated using different norm constraints: $L_1$, $L_2$, and $L_\infty$, as well as results from PhotoGuard and DiffusionGuard.}
\label{fig:imm-attack-2}
\end{figure*}


\newpage
\subsection{Prompt-Agnostic Immunization Experiment}
\label{sec:prompt_agonstic_noise}

\begin{figure*}[ht]
\centering
\includegraphics[width=0.5\columnwidth]{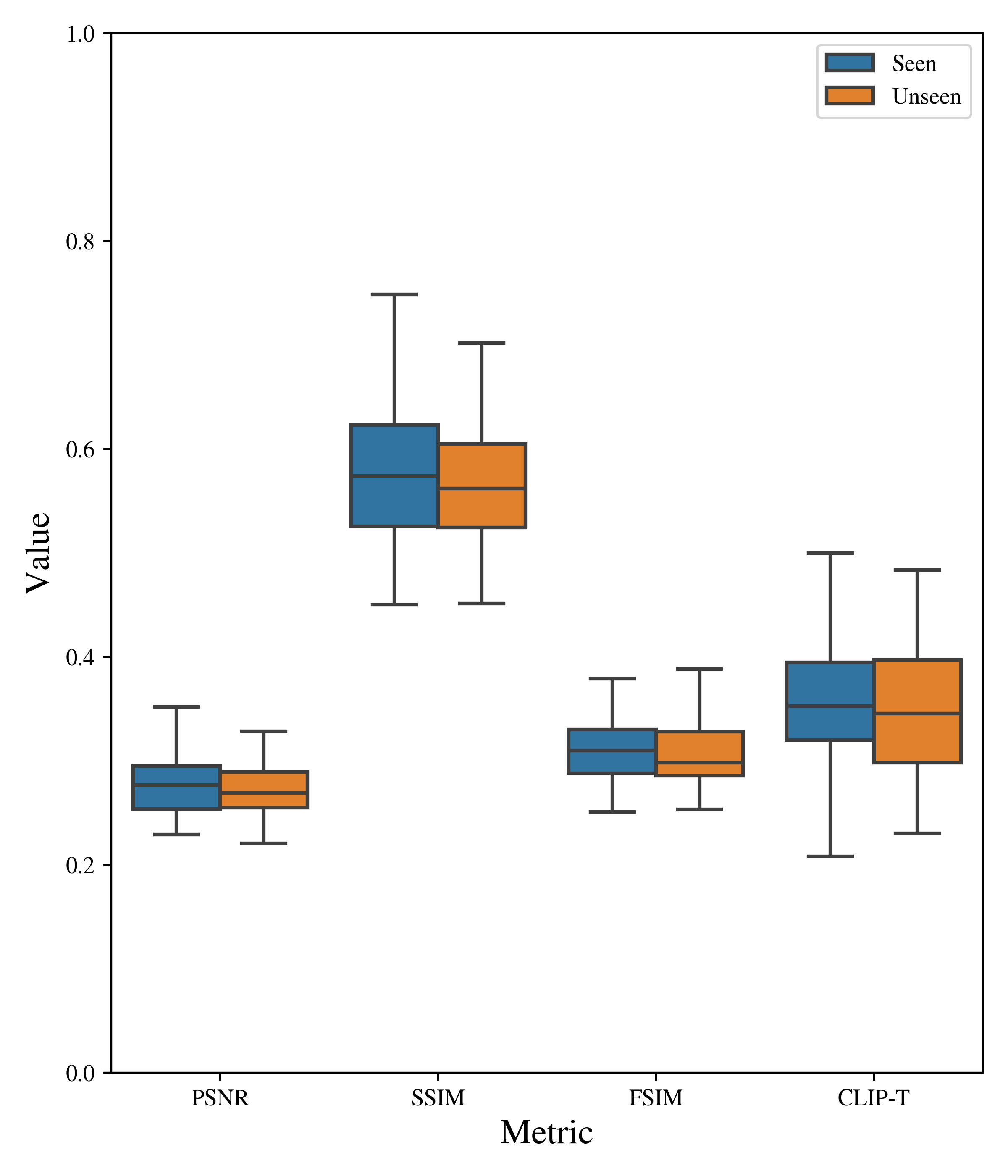}
\caption{\textit{\textbf{Experiment results for prompt-agnostic noise.}} We present our performance metrics between prompts for 75 prompts seen in training (blue color) and 75 prompts unseen in training (orange color). PSNR and CLIP-T values are divided by 50 for visualization purposes. We can see that the two distributions are almost identical, suggesting that our method performs similarly across all prompts, suggesting the prompt-agnostic nature of our \methodname.}
\label{fig:prompt_agnostic_noise_fig}
\end{figure*}

We conduct additional experiments to demonstrate that the noise produced by our \methodname\ (and consequently the immunized images) is prompt-agnostic. To achieve this, we train \methodname\ three times, using a different image for each training setup. In each experiment, we use a single image with 100 seen prompts for training and evaluate it on 75 seen prompts and 75 unseen prompts (not included in the training set). The results are then averaged across all images for each prompt. As shown in Fig.~\ref{fig:prompt_agnostic_noise_fig}, the quantitative results for seen and unseen metrics are highly similar, and the low variances further confirm that the noise generalizes effectively across diverse prompt conditions.

\newpage
\subsection{Loss Weight Selection}
\label{sec:complementary_ablation_study}

The hyperparameter $\alpha$ in \methodname's loss function controls the balance between imperceptibility and edit disruption. It is defined in the overall loss as $\mathcal{L} = \alpha \cdot \mathcal{L}_{\text{noise}} + \mathcal{L}_{\text{edit}}$, where a larger $\alpha$ emphasizes minimizing visible noise, potentially at the cost of reduced editing resistance, and a smaller $\alpha$ enhances robustness to edits but may introduce more perceptible perturbations.

To determine an optimal value for $\alpha$, we conduct an ablation study on a subset of 100 images, evaluating three values: $\alpha = 2$, $4$, and $6$. The results are summarized in Table~\ref{tab:alpha_ablation}. We observe that while increasing $\alpha$ improves imperceptibility—as indicated by slightly higher SSIM (Noise) and PSNR scores—the edit disruption becomes weaker, reflected in a deterioration of the SSIM and PSNR metrics.

We select $\alpha = 4$ as the optimal configuration. It provides a strong balance between imperceptibility and disruption: the gain in SSIM (Noise) from $\alpha = 4$ to $\alpha = 6$ is marginal, while the drop in editing robustness is more pronounced. Furthermore, qualitative inspection confirms that the perturbations at $\alpha = 4$ are already imperceptible, making further increase in $\alpha$ unnecessary.

\begin{table}[ht]
\centering
\caption{\small{\textit{\textbf{Ablation study on the loss weight $\alpha$ in $\mathcal{L} = \alpha \cdot \mathcal{L}_{\text{noise}} + \mathcal{L}_{\text{edit}}$.}} Metrics demonstrate the trade-off between imperceptibility and edit disruption. Best values for SSIM (Noise) are bolded, while lower SSIM and PSNR indicate stronger editing disruption.}}
\begin{tabular}{lccc}
\toprule
\textbf{Configuration} & SSIM $\downarrow$ & PSNR $\downarrow$ & SSIM (Noise) $\uparrow$ \\
\midrule
\methodname\ w/ $\alpha = 2$ & \textbf{0.536} & \textbf{14.47} & 0.987 \\
\methodname\ w/ $\alpha = 4$ & 0.588 & 15.38 & \textbf{0.993} \\
\methodname\ w/ $\alpha = 6$ & 0.625 & 16.23 & \textbf{0.996} \\
\bottomrule
\end{tabular}%
\label{tab:alpha_ablation}
\end{table}

\newpage
\subsection{Limitations}
\label{sec:limitations}

\begin{figure*}[ht]
\centering
\includegraphics[width=\textwidth]{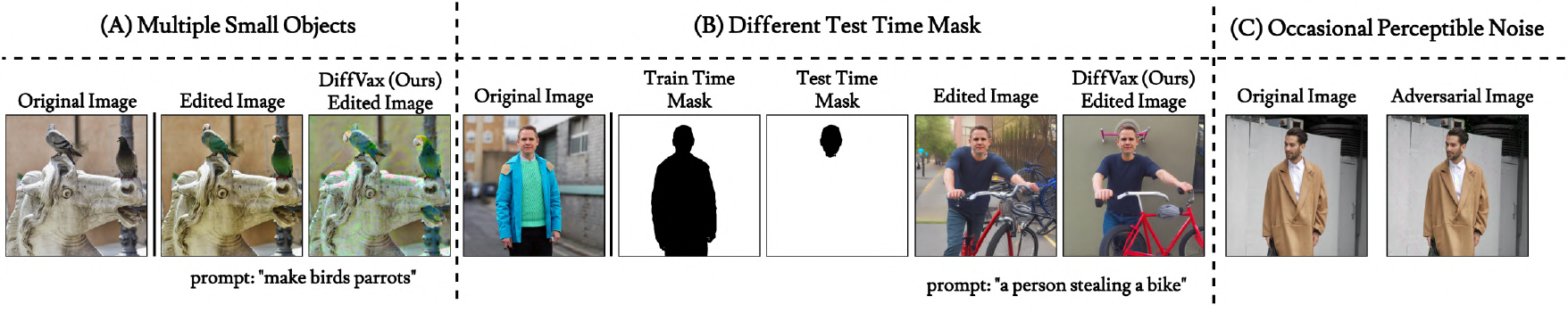}
\caption{\textit{\textbf{Limitations.}} We highlight three failure cases: (A) \methodname\ struggles to protect multiple small objects simultaneously; (B) protection may fail when the test-time editing mask (e.g., a small face region) drastically differs from the training-time immunization mask (e.g., full body); and (C) in rare instances, the immunization noise becomes perceptible to the human eye.}
\label{fig:limit}
\end{figure*}

Despite its strong performance, \methodname\ exhibits certain limitations, as illustrated in Figure~\ref{fig:limit}. First, the model faces challenges in effectively immunizing scenes with multiple small objects (Panel A). In such cases, the protective noise may be too dispersed to disrupt the semantic guidance for every individual object, allowing some edits (e.g., turning birds into parrots) to succeed. Second, while generally robust to mask variations, extreme discrepancies between the immunization and editing masks can compromise protection. For instance, if an image is immunized with a full-body mask but the attacker employs a significantly smaller, localized mask (Panel B), the edit may partially bypass the immunization. Finally, although we optimize for imperceptibility, occasional perceptible artifacts may appear (Panel C), particularly in smooth or uniform regions where the adversarial perturbations are harder to conceal.

\subsection{Future Directions}
While DiffVax demonstrates strong generalization within the latent diffusion paradigm, a critical avenue for future research is achieving universal cross-architecture immunization. Current perturbations are optimized for the gradient flow of U-Net-based LDMs, which may not fully transfer to emerging non-LDM architectures, such as flow-matching models or pixel-space diffusion transformers. We envision extending our scalable feed-forward framework to incorporate \textit{Ensemble Adversarial Training}, where the immunizer is trained simultaneously against a diverse set of backbones. This would encourage the model to learn universally disruptive features that transcend specific architectural biases. Additionally, expanding the training distribution to include non-photorealistic domains, such as anime and digital art, will be essential for broader real-world applicability.

Furthermore, while our method effectively immunizes video via per-frame processing, future iterations could explicitly model temporal dynamics. Integrating temporal consistency objectives directly into the training loop would allow the immunizer to exploit motion priors, potentially creating ``video-native'' perturbations that are even more robust to temporal filtering and video compression. Finally, addressing the physical limitations of extremely small editing masks remains an open challenge; exploring frequency-domain regularization or perceptual loss functions that adaptively balance noise visibility with protective strength in constrained regions could offer a pathway to more granular control.

\newpage

\subsection{Reproducibility Statement}
\label{a:reproducibility}
The source code of the project is provided in the supplementary. Project can be reproduced by following the provided guidelines and source code. All experiments can be replicated using the instructions and datasets referenced in this paper. 

\subsection{Ethics Statement}
\label{a:ethics}
This work does not raise any foreseeable ethical concerns. The experiments were conducted solely on publicly available datasets. 

\subsection{LLM Usage Statement}
\label{a:llm}
Large language models (LLMs) were used exclusively for assistance in grammar correction, formatting, and improving the clarity of writing. They were not employed for generating research ideas, designing experiments, or creating results.

\end{document}